\documentclass[review]{elsarticle}
\usepackage[UKenglish]{babel}
\usepackage[utf8]{inputenc}
\usepackage{xcolor}
\usepackage{graphicx}
\usepackage[a4paper, total={7in, 10in}]{geometry}

\usepackage{helvet}
\usepackage[hidelinks]{hyperref}

\usepackage{fullpage}
\usepackage{lscape}
\usepackage{pdfpages}
\usepackage{multirow}
\usepackage{afterpage}
\usepackage{amsmath}
\usepackage{array}
\usepackage{tabularx}
\usepackage{longtable}
\usepackage[para]{footmisc}
\usepackage{multicol}

\newcommand{\x}[0]{\mathbf{x}}

\begin{document}
\begin{frontmatter}

\title{
CACTUS as a Reliable Tool for Early Classification of Age-related Macular Degeneration
}
\author[1,2]{Luca Gherardini\corref{cor1}}
\ead{l.gherardini@sanoscience.org}
\author[2]{Imre Lengyel}
\author[3]{Tunde Peto}
\author[4,5,6,7]{Caroline C.W. Klaver}
\author[4,5]{Magda A. Meester-Smoor}
\author[4,5]{Johanna Maria Colijn}
\author{EYE-RISK Consortium}
\author{E3 Consortium}
\author[1,3]{Jose Sousa}

\cortext[cor1]{Corresponding author}
\affiliation[1]{
    organization={Personal Health Data Science Team, Sano Centre for Computational Personalised Medicine},
    city={Krakow},
    country={Poland}
}

\affiliation[2]{
    organization={The Wellcome-Wolfson Institute for Experimental Medicine, School of Medicine Dentistry and Biomedical Science, Queen's University Belfast},
    city={Belfast},
    state={Northern Ireland},
    country={the United Kingdom}
}

\affiliation[3]{
    organization={Institute of Clinical Sciences Building A, Queen’s University Belfast},
    city={Belfast},
    state={Northern Ireland},
    country={the United Kingdom}
}

\affiliation[4]{
    organization={Department of Epidemiology, Erasmus Medical Center},
    city={Rotterdam},
    country={the Netherlands}
}

\affiliation[5]{
    organization={Department of Ophthalmology, Erasmus Medical Center},
    city={Rotterdam},
    country={the Netherlands}
}

\affiliation[6]{
    organization={Department of Ophthalmology, Radboud University Medical Centre},
    city={Nijmegen},
    country={the Netherlands}
}

\affiliation[7]{
    organization={Institute of Molecular and Clinical Ophthalmology Basel},
    city={Basel},
    country={Switzerland}
}

\begin{abstract}
Machine Learning (ML) is widely used to solve various tasks, including disease classification, monitoring, and prediction.
The performance of ML models is subject to the availability of significant volumes of complete data.
Unfortunately, the data available in real healthcare scenarios is often severely limited or incomplete, impairing the performance of these ML models in practice.
While these already limit performance, other problems may also emerge, such as the trustworthiness of solutions depending on the datasets. 
These limitations, combined with the opacity of some ML models, can make understanding their behaviour and adoption difficult.
There is a push for transparent and easily understandable artificial intelligence in several fields, especially healthcare.
Age-related Macular Degeneration (AMD) is a retinal disease affecting millions of older adults worldwide.
The lack of effective treatments to reverse its progression makes early diagnosis crucial to adopt preventive strategies and identify those at risk of losing sight.
Diagnosing AMD mainly relies on assessing retinal images in addition to verbal reports of symptoms by the patients. 
There is a need for a different classification approach focused on genetic, dietary, clinical, and demographic factors. 
Recently, we developed a new model called ``Comprehensive Abstraction and Classification Tool for Uncovering Structures'' (CACTUS) to understand how to classify AMD stages.
We propose an application of CACTUS to tackle these issues and provide an explainable and flexible tool for early AMD classification.
CACTUS outperforms standard Machine Learning models, helps identify key decision-making factors, and provides confidence in its outcomes to improve decision-making.
The features highlighted by CACTUS as the most important for its reasoning allowed us to compare and test them against existing medical knowledge.
By removing less relevant or biased information from the dataset, we simulated a clinical scenario involving clinicians who can provide feedback and identify biases.
\end{abstract}


\end{frontmatter}

\section*{Introduction}
Diagnosing a disease requires dealing with the intrinsic uncertainty of the many parameters available and their interdependence.
Artificial Intelligence (AI), and in particular, some of its most relevant sub-categories, such as Machine Learning (ML) and Deep Learning (DL), are widely adopted to address these uncertainties, given that enough data and computational power are provided~\cite{Adadi2021}.
However, these requirements are not always feasible.
In fields like healthcare, for example, collecting more data might not be possible due to practical reasons (e.g., economic costs, required equipment and staff, and the rarity of a disease)~\cite{Lee2022}.
Another typical property of medical data is the high percentage of missing values and the presence of noise, representing a substantial obstacle for current AI systems~\cite{Wang2019}.
Using DL as a universal learning model is challenging due to its intrinsic lack of transparency, commonly called ``black box''~\cite{Handelman2019, London2019, Das2020, Angelov2021}.
This opacity overshadows its practical effectiveness in modelling and predicting complex patterns, as its internal reasoning is, by default, complex to validate.
As such, explaining DL models is challenging, making their adoption in sensitive fields like law, healthcare, and economics debatable~\cite{Miotto2017}.
Despite recent efforts to make DL models more explainable~\cite{Adamson2018}, it is not easy to overcome their biases and excessive sensitivity towards the input~\cite{Du2021}.
More explainable models are also easier to \emph{trust} and rely on, which is crucial for integrating AI systems into our society.
Imbalances in the dataset can, for example, lead to biases in the classification~\cite{Wang2019, Adamson2018}.
The increasing awareness of the role of AI in our daily lives led to regulations to define the liability of an algorithm: the European Union integrated the ``right to explainability'' in recital 71 of the General Data Protection Regulation (GDPR).
At the same time, the US Congress decreed the ``Algorithmic Accountability Act'' (HR 6580 of 2022)~\cite{MacCarthy2020, GoodmanBryce2017}.
Interpretable models are essential to understanding decision-making and improving their judgment by detecting malfunctions and biases~\cite{Guidotti2021}.
Therefore, research is gradually shifting towards increasing the explainability of existing DL models (ad-hoc and post-hoc solutions) or testing altogether different algorithms~\cite{Das2020, Li2022}.
Users, such as clinicians and patients, must understand the decisions that AI makes.
Evidence suggests that knowledge graphs~\cite{TIDDI2022103627} are particularly suitable for deploying ``explainability'' and supporting reasoning~\cite{Booch2010}.
They were also used in the biomedical sector to describe concepts, events and their interdependence~\cite{NICHOLSON20201414}.

Age-related macular degeneration (AMD) is a chronic retinal disease that accounts for $8.7\%$ of worldwide blindness~\cite{Wong2014}.
The number of individuals affected by AMD is expected to increase from $196$ million reported in 2020 to $288$ million by 2040~\cite{Wong2014}.
Its development is affected by environmental, dietary, genetic, and lifestyle factors, but data for these factors are often patchy in population-based study datasets.
Overall, AMD has two advanced forms: geographic atrophy (GA) and choroidal neovascularisation (CNV); currently, outside of the USA~\cite{Spiewak2024}, only the latter has available treatment.
These are expensive and very demanding for patients, carers, and society due to comorbidities, patient discomfort and, in some cases, diminishing effects or futility~\cite{Thomas2022}.
The progression of AMD is slow but irreversible.
Therefore, preventive strategies are needed to delay its onset of end-stage disease.
Lifestyle and diet changes may reduce the speed and likelihood of progression to later AMD stages~\cite{Chew2020}.
We, therefore, need to identify and then focus on developing earlier diagnoses to improve the quality of life of people affected by AMD.
AMD is usually diagnosed through retinal imaging and image analysis.
This diagnosis approach requires trained specialists, specific equipment, and time, and it relies on human interpretation of the images~\cite{Liu2019}. These elements, combined with increasing demand, are exceeding the capacity of available specialists~\cite{Liu2019}.
Given the limited treatment options and the potential for preventive strategies, early diagnosis is a significant unmet medical need.

The Comprehensive Abstraction and Classification Tool for Uncovering Structures (CACTUS)~\cite{Gherardini2024} analyses and \emph{abstracts} data to build knowledge graphs~\cite{GARNELO201917}, representing potential interaction between different features.
CACTUS can handle missing values and noisy data by leveraging novel approaches~\cite{Gherardini2024}.
To manage the high complexity of AMD classification, CACTUS needed to undergo significant changes and expansion since its last publication to improve its generalisation capabilities and transparency. 

\section*{Results}

\subsection*{CACTUS performance}
\begin{figure}[h!]
    \centering
    \includegraphics[width=.8\textwidth]{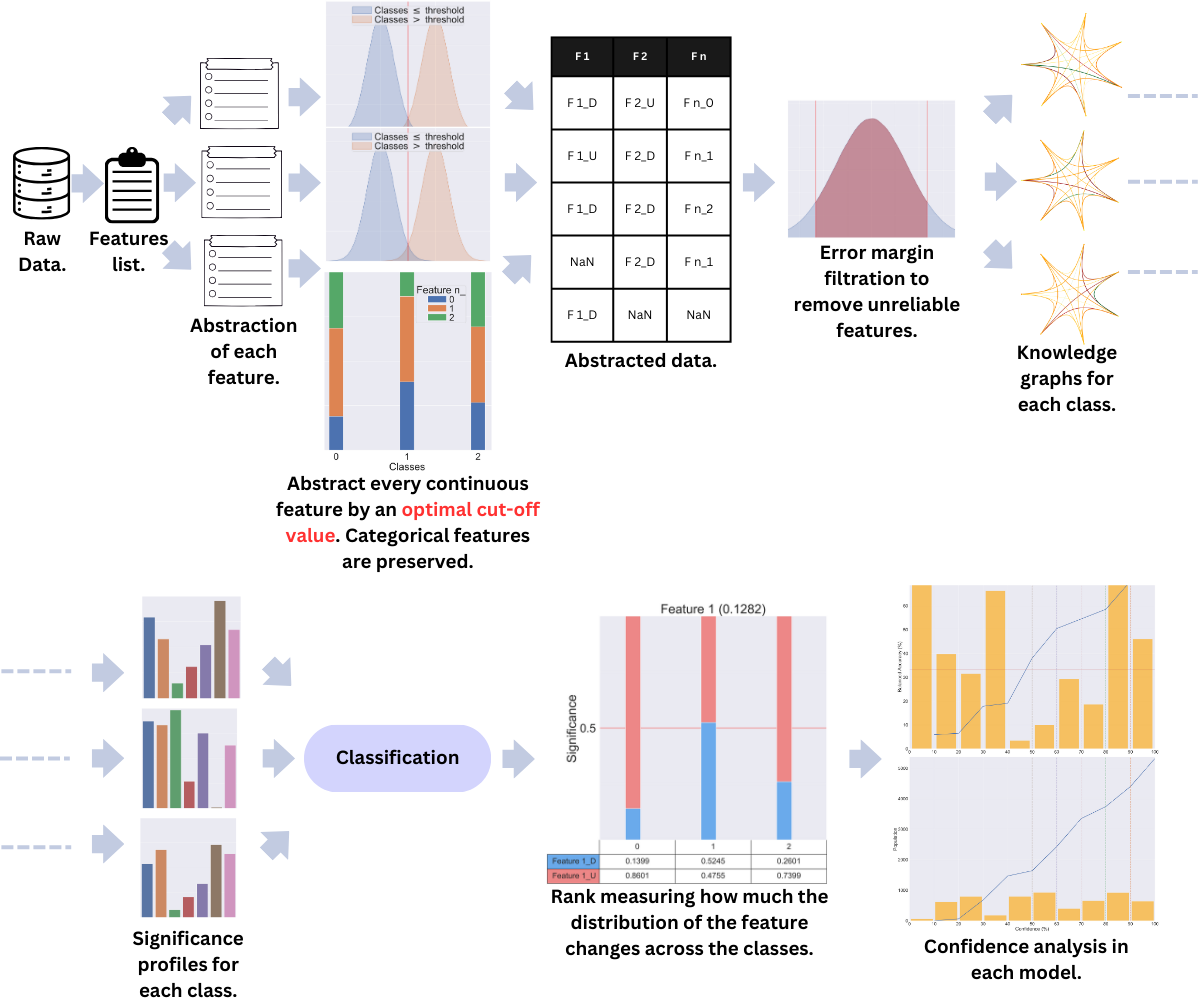}
    \caption{
    \textbf{A diagram of the internal functioning of CACTUS}.
    Each continuous feature in the dataset is partitioned into \emph{flips} representing high (\_U) or low (\_D).
    Otherwise, it is kept unaltered.
    The threshold to partition the continuous features is found using a receiver operating characteristic (ROC) curve.
    Two populations are required to build the ROC curve and find the most appropriate threshold to abstract the features.
    These are created by partitioning the available AMD classes (e.g., by considering the healthy individuals as one population and all the AMD groups as another) into two groups.
    In previous versions, the user had to select a universal threshold to divide the classes, which required prior domain knowledge and multiple comparisons.
    Still, CACTUS has since been improved to select the most appropriate threshold for each feature, which is a more sensitive approach to consider the uniqueness of each feature in distinguishing between a certain set of classes.
    Discrete values allow for building a knowledge graph representing each class by linking the elements by their conditional probability.
    The centralities computed on these represent a Significance profile.
    Thus, the classification process compares each individual against the available representations to assign the most similar one.
    The classification is then applied to all the available individuals to assess the performance.
    The significance of each flip in each class is analysed to identify the elements driving the classification and provide an explanation for the behaviour of the model.
    }
    \label{fig:pipeline}
\end{figure}

\begin{figure}[h!]
    \centering
    \includegraphics[trim={0 0 0 1cm},clip,width=.5\textwidth]{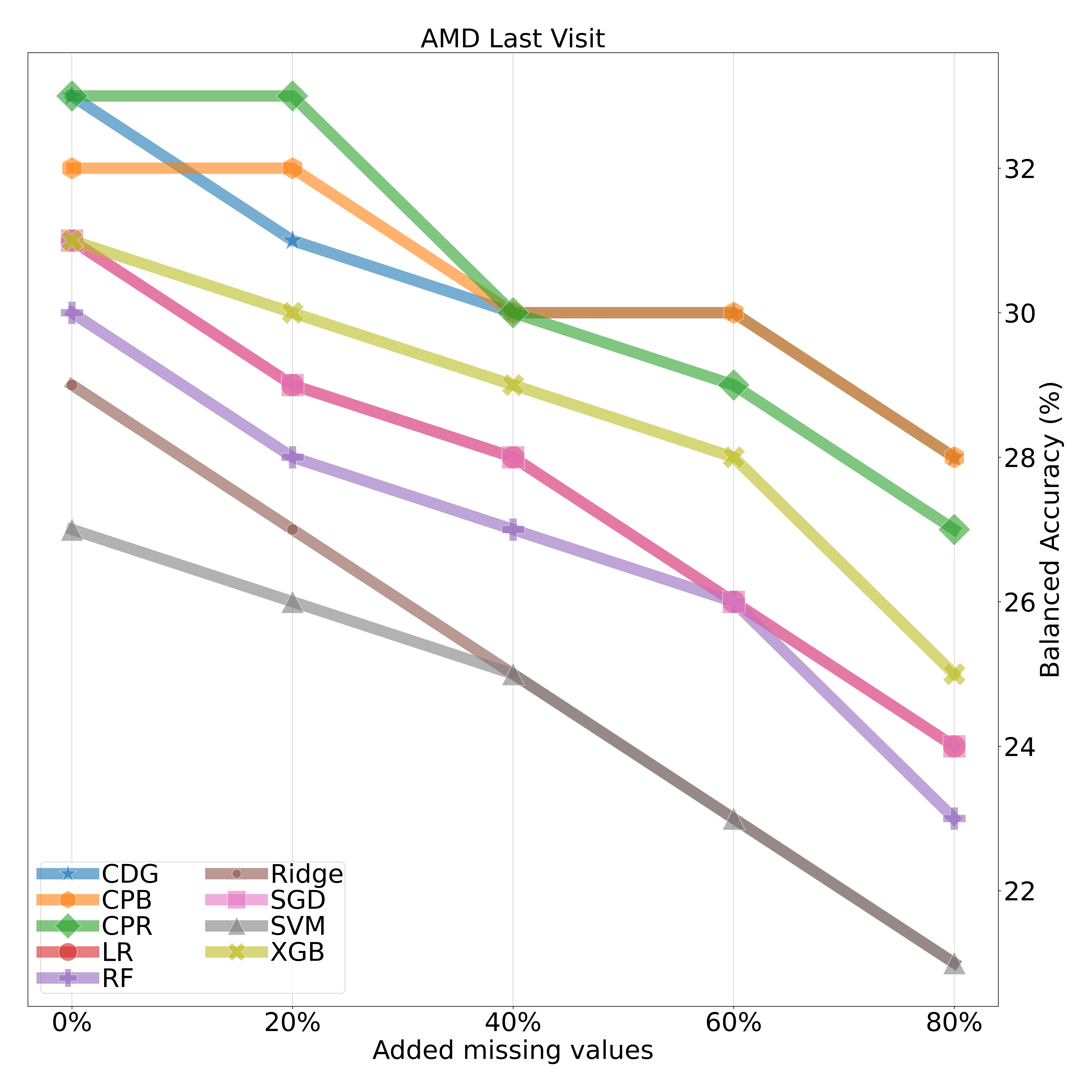}
    \caption{
        \textbf{The performance achieved by CACTUS compared with ML models}.
        CACTUS's Degree (CDG, blue stars), PageRank (CPR, green diamonds), and Probabilistic (CPB, orange hexagons) implementations are shown along standard ML models: Linear Regression (LR, red circles), Random Forest (RF, purple plus signs), Ridge (brown dots), Stochastic Gradient Descent (SGD, pink squares), Support Vector Machine (SVM, grey triangles), and eXtreme Gradient Boosting (XGB, yellow Xs).
        X-axis: the percentage of values that were removed from the dataset. 
        Y-axis: the balanced accuracy achieved by the algorithms in distinguishing between AMD stages.
        $20\%$ represents the random chance of correctly guessing the right AMD stage.
    }
    \label{fig:graphical-comparison}
\end{figure}

First, we compared CACTUS against standard ML models for different levels of removed values.
For this, we followed our workflow, depicted in Figure~\ref{fig:pipeline} and described in the Methods section.
To compare these models, we relied on balanced accuracy, which accounts for imbalances between the groups, as described in Table~\ref{tab:dataset-stages}.
The results, summarised in Table~\ref{tab:classification-results}, indicate that CACTUS could better differentiate between multiple AMD stages at every level of fragmentation.
All algorithms achieved more than the random chance ($20\%$ or 1 out of 5) of correctly guessing the right AMD stage.
The relationship between the models' balanced accuracy (y-axis) and the percentage of removed values (x-axis) is depicted in Figure~\ref{fig:graphical-comparison}.
The plot shows higher performance of all three CACTUS classification methods (Degree, CDG; Probabilistic, CPB; and PageRank, CPR) compared to traditional ML algorithms: Ridge, Random Forest (RF), Logistic Regression (LR), Stochastic Gradient Descent (SGD), Support Vector Machine (SVM), and eXtreme Gradient Boosting (XGB).
PageRank (CPR) showed the highest balanced accuracy in the first three versions of the dataset ($0\%$, $20\%$, and $40\%$ of missing values), while Degree (CDG) and Probabilistic (CPB) achieved the highest in the 3 most fragmented ($40\%$, $60\%$, and $80\%$).
The performance degradation due to fragmentation ranged from $4\%$ (CPB) to $8\%$ (Ridge) loss in balanced accuracy from the most complete to the most fragmented dataset in all models.

\subsection*{CACTUS confidence}
\begin{figure}[h!]
    \centering
    \includegraphics[width=.8\textwidth]{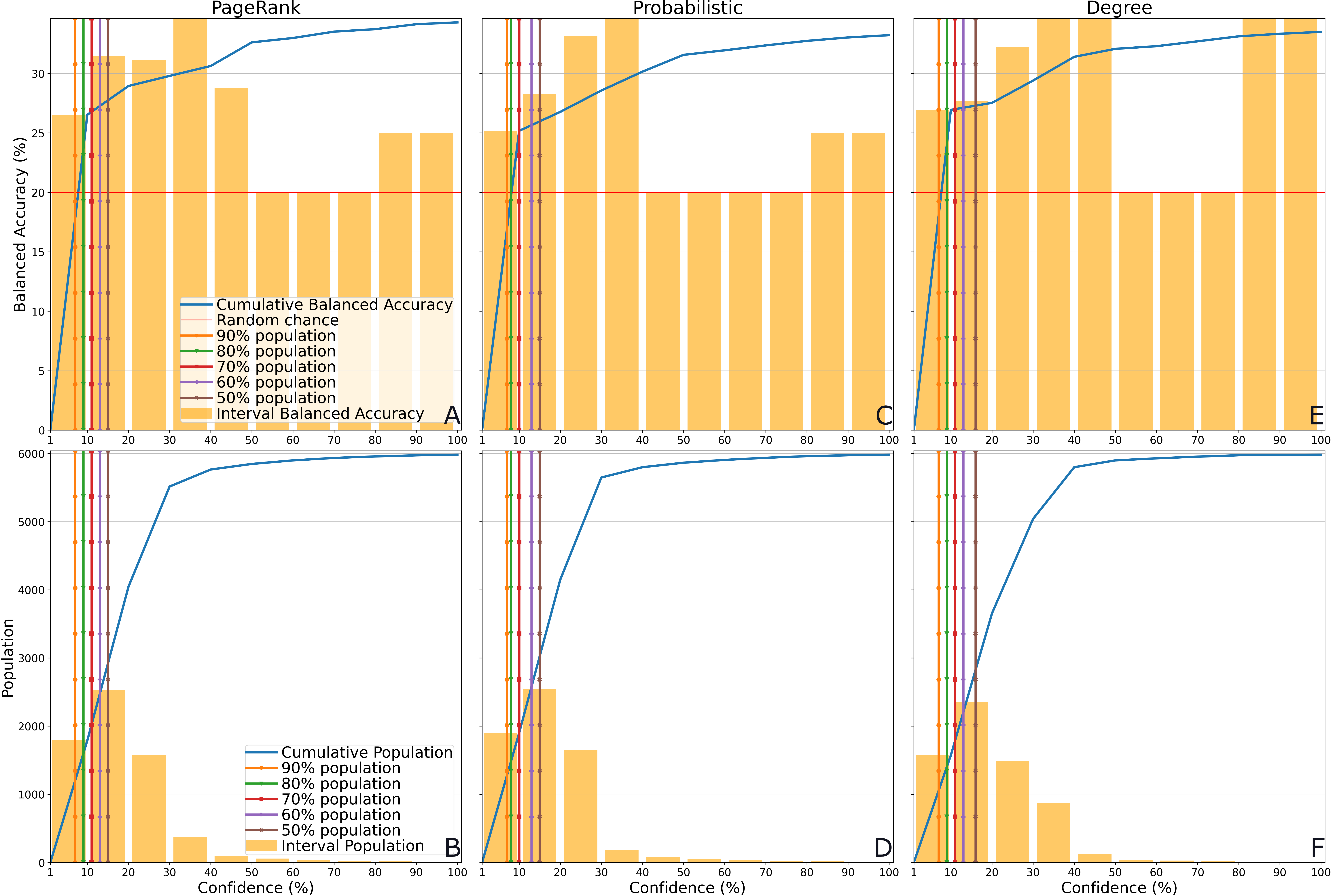}
    \caption{
        \textbf{The relationship between confidence (X-axis), balanced accuracy (Y-axis, first row), and population (Y-axis, second row)}.
        Confidence is reported on the X-axis, while balanced accuracy and population are reported on the Y-axis of, respectively, the first and second rows.
        This metric is provided for the Degree (A, B), PageRank (C, D), and Probabilistic (E, F) classification methods in CACTUS.
        The confidence (blue line) is computed as the similarity of a sample to the most similar class compared to its closeness to the others.
        The yellow bins (bars) represent the values (the balanced accuracy in panels A, C, and E; the number of people in panels B, D, F) corresponding to a confidence interval (e.g., the performance of CPR when it has a confidence between 20\% and 30\%).
        The population is expressed as the number of people for which a given algorithm had a certain confidence level.
        Cumulative values (blue lines) represent the balanced accuracy/population when incrementally including all the intervals (yellow bins) on their left side (e.g., the cumulative balanced accuracy at $20\%$ is the average of the balanced accuracy in the $[0, 10), [10, 20]$ bins).
        The cumulative balanced accuracy is weighted by the number of people in the corresponding bins in the lower plots).
        The vertical lines represent the confidence levels to cover 90\%, 80\%, 70\%, 60\%, and 50\% of the population. 
        The horizontal orange line crossing panels A, C, and E represents the random balanced accuracy, which can be used as another indicator for comparing the quality of the models.
    }
    \label{fig:confidence}
\end{figure}

Figure~\ref{fig:confidence} describes the relationship between the confidence of CPB, CPR, and CDG in assigning AMD stages, their balanced accuracy, and the population.
This relationship can set confidence thresholds covering 90\%, 80\%, 70\%, 60\%, and 50\% of the population, as well as the corresponding classification performance, represented by vertical coloured lines in Figure~\ref{fig:confidence}. 
The more inflated the cumulative lines for confidence and populations, the better the corresponding model.
This indicates that the model performs better while requiring lower confidence. 
This shows that, despite the good resistance of CPB and CDG to missing values, their confidence and population distributions are skewed to high values (CPB, panels C and D; CDG, panels E and F), meaning that they can be trusted only when they return high confidence levels.
Conversely, CPR has a more normally distributed confidence on the population and more inflated cumulative curves (panels A and B), allowing it to provide good coverage of the population with reasonable confidence.
For these reasons, we omitted CPB and CDG from further evaluations and discussions, focusing only on CPR.
For instance, a confidence threshold of $15\%$ enables us to confidently apply CPR on $50\%$ of the population (brown vertical line), with a balanced accuracy of $27\%$, which is above the random chance line ($20\%$; orange horizontal line).

\subsection*{CACTUS ranks}
\begin{figure}[h!]
	\centering
	\includegraphics[width=.8\textwidth, height=\textheight, keepaspectratio]{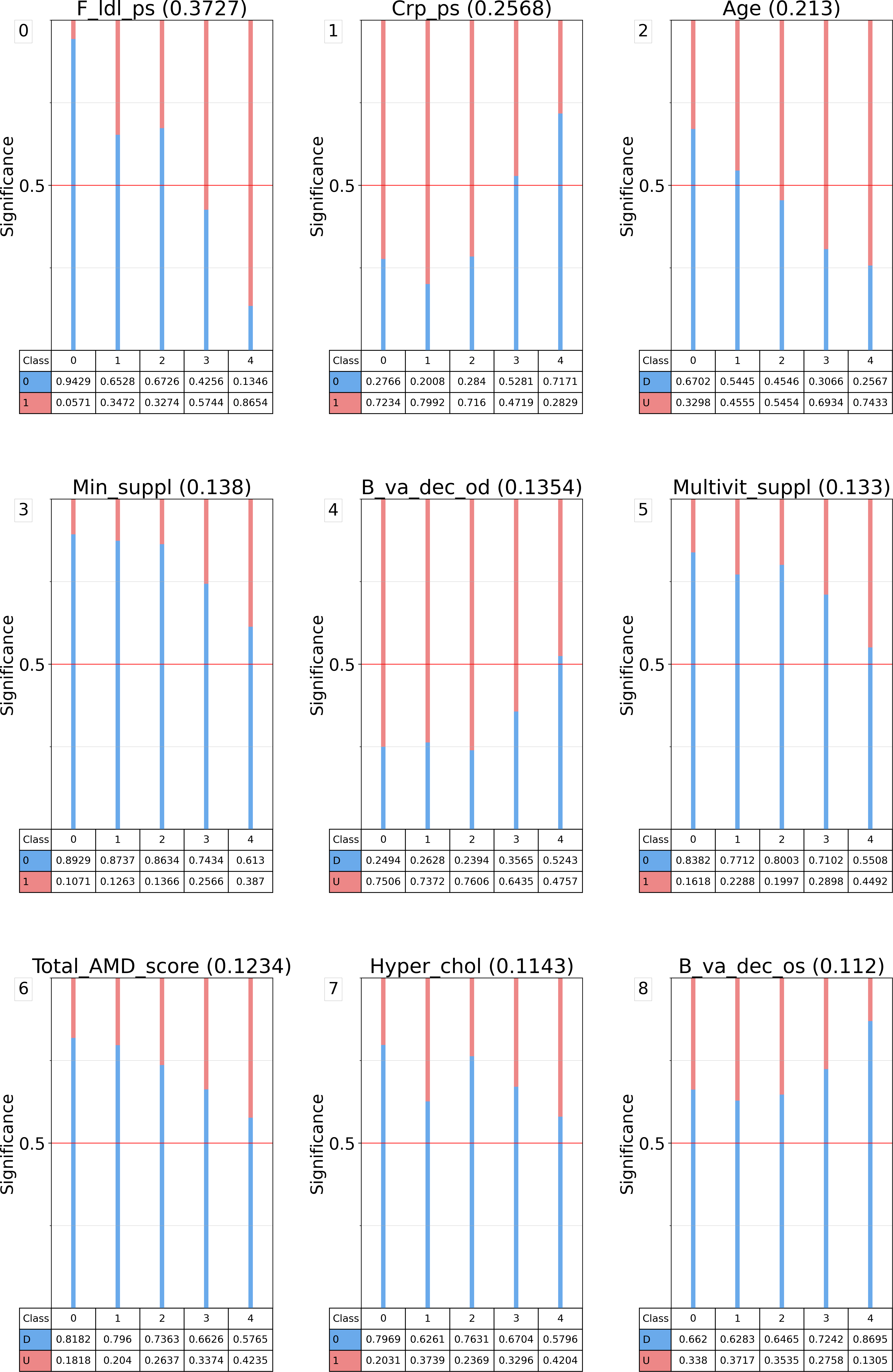}
	\caption{
		The ranks of the features computed by CACTUS through the \emph{PageRank} centrality on the knowledge graphs.
		They are sorted from the highest (upper left) to the lowest (lower right), as indicated in the title of each plot between parentheses.
	}
	\label{fig:ranks-pagerank}
\end{figure}

After identifying PageRank (CPR) as the most accurate model, we determined the ranks of the nine most influential features in this model (Figure~\ref{fig:ranks-pagerank}).
The ranks also describe the distribution of the flips across the classes to show why a feature is important.
Flips are values abstracted by CACTUS by finding an optimal threshold to differentiate between AMD stages. 
Up (U) and Down (D) values represent values above and below this optimal threshold. 
SNPs, such as ARMS2\_rs3750846, have more than two flips representing the three possible alleles. 
A total list of the ranks returned by CPR, CDG and CPB is available in the supplementary material (Supplementary Table~\ref{tab:ranks}).
For presentation purposes, the significance of the abstracted values of the same feature was normalised to sum up to 1 in each class.
For CPR, the significance of an abstracted value combines its probability of appearing in a given AMD stage with its centrality in the corresponding graph structure.
By analysing the clinical features, we found that their significance changes according to the disease stages, while others show a variable dynamic.

\begin{figure}[h!]
	\centering
	\includegraphics[width=.8\textwidth, height=\textheight, keepaspectratio]{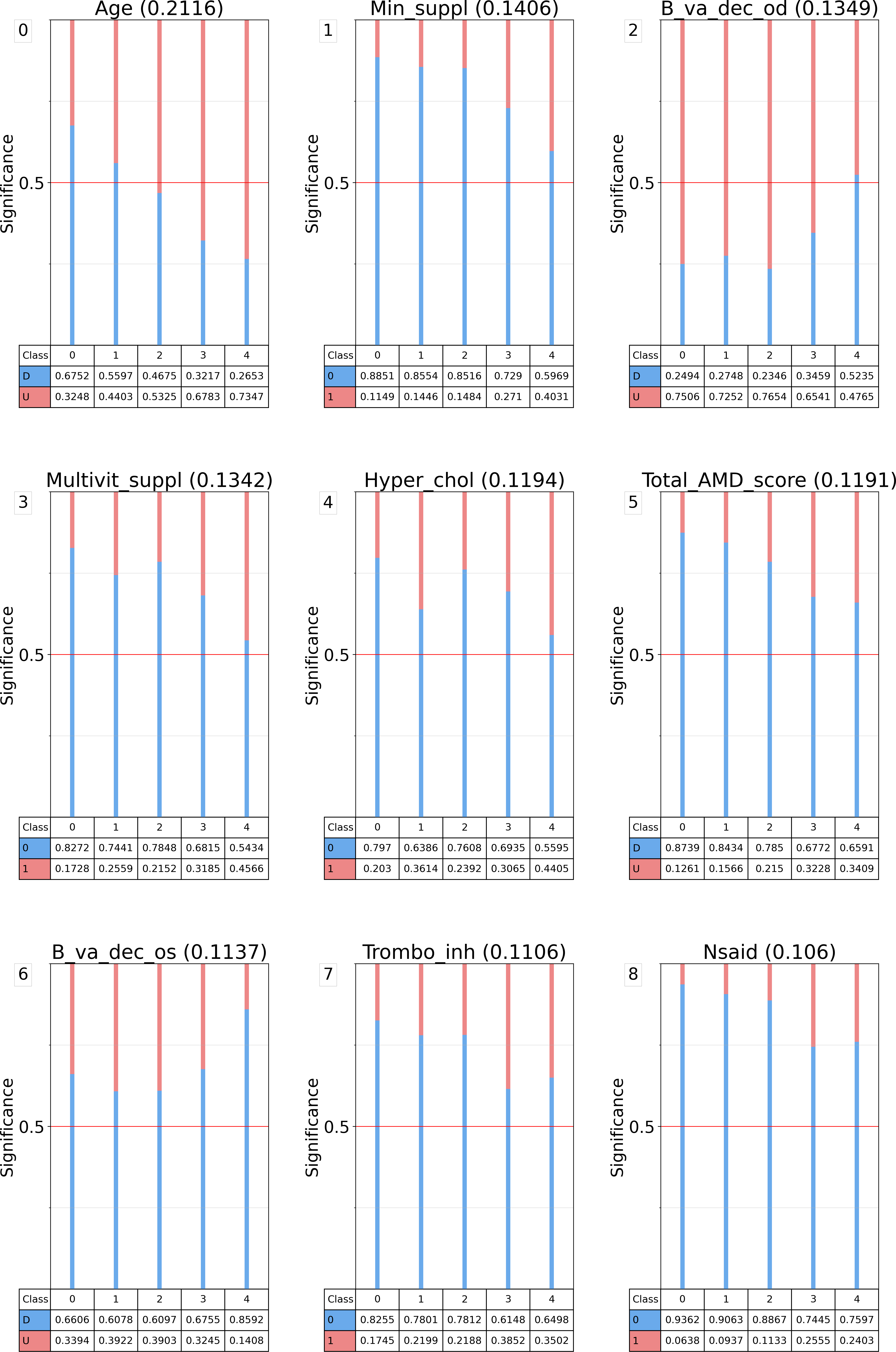}
	\caption{
		\textbf{The ranks of the 9 most important features for the \emph{PageRank} metric computed on the filtered dataset}.
		The most relevant features are sorted from the highest (upper left, 0) to the lowest (lower right, 8), as indicated in the title of each plot between parentheses.
		The distribution of each abstracted value (Down in blue and Up in pink) for each AMD stage (0, 1, 2, 3, 4) is shown.
        Up (U) and Down (D) represent values above and below the optimal threshold found by CACTUS during the abstraction process.
	}
	\label{fig:corrected-ranks}
\end{figure}

Furthermore, we noted that features unrelated to AMD were part of the dataset, for example, those describing the type of tissues collected from the patient.
We gave the model feedback and refined the feature pool by excluding those without biological significance for AMD and those with more than 50\% missing values.
The choice for this threshold reflected the population threshold used to compare CPR, CDG and CPB.
After removing these features and rerunning the model, we obtained a new set of ranks, shown in Figure~\ref{fig:corrected-ranks}, revealing the underlying biology of AMD.
Increasing Age (rank 0), mineral and multivitaminic supplements (ranks 1 and 3, respectively), hypercholesterolemia (rank 4), total AMD genetic score (rank 5), and the assumption of anti-trombotic agents (trombo\_inh, rank 7), were associated with more advanced disease stages, indicating that older people with a higher genetic predisposition and comorbidities related to AMD are more likely to develop advanced stages.
Indeed, this also reveals the obvious information that people with advanced AMD are more inclined to take supplements aimed at slowing its progression.
Decreases in best-corrected visual acuity (B\_va\_dec\_os and B\_va\_dec\_od) become a significant determinant at the late stages of the disease.
This filtration decreased the balanced accuracy by 0\% to 2\% for CDG, CPR, and CPB, as shown in Table~\ref{tab:ba-comparison}.

Lastly, we ran the model on the nine highest ranks identified in Figure~\ref{fig:corrected-ranks} to measure their contribution to the classification task.
Table~\ref{tab:ba-comparison} shows that these nine features alone account for most of the model's performance, as the balanced accuracy lost, at most, $5\%$.
Interestingly, CDG achieved a higher accuracy on the 9 highest ranks alone than on the refined dataset, surpassing the other two algorithms.

\section*{Discussion}
The classification performance of CACTUS highlighted how our algorithm can provide higher balanced accuracy than standard ML models.
CACTUS is a useful tool for practical clinical scenarios, as the balanced accuracy quantifies how well an algorithm is expected to perform outside the experimental setting. 
The random chance to guess the right AMD stage was $20\%$ (1 in 5) in our analysis.
Algorithms that perform poorly in controlled testing environments cannot provide reliable performance in real-life applications.
The balanced accuracy achieved by each CACTUS method for each version of the dataset is well above this threshold and the other ML methods, showing that the fragmented clinical data used can still provide an informative classification for AMD, and the algorithms were able to extract meaningful patterns to aid a clinician. 
This was the first time CACTUS was compared with XGBoost, which natively deals with missing values.
CACTUS PageRank (CPR) achieved the highest balanced accuracy in 3 out of 5 experiments, as shown in Figure~\ref{fig:graphical-comparison}, and proved the most solid confidence for applications in real-case scenarios, as displayed in Figure~\ref{fig:confidence}.
For these reasons, we focused on CPR to analyse its performance in AMD classification.
When we refined the dataset to exclude information without biological and clinical relevance for AMD or simply too scarce to be reliable for its classification, we emulated the kind of feedback CACTUS could receive from a clinician in a real-life application.
This produced refined ranks (Figure~\ref{fig:corrected-ranks}) that displayed the interactions inside each AMD stage and how they may change.
Therefore, CPR may offer a new perspective on AMD characterisation and classification that outperforms other ML algorithms.

The expansion of CACTUS included a confidence metric for practical clinical scenario approaches, a better abstraction mechanism and graph modelling, and a redesigned classification function.
The main advantage of having a measure of confidence is to identify if more data or different expertise is needed to overcome the \emph{knowledge limits}.
CACTUS operates based on the available data but can indicate if having more information would make the outcome more reliable.
The confidence threshold for accepting the model's decision lacks a formal and objective definition but undoubtedly increases the transparency of the tool and its ease of use.
Analogously to probabilities, there is no objective definition of ``enough'' that can be obtained a priori, but it rather depends on subjective criteria, such as the allowed risk, trust in the system, resources required for collecting more data, and/or availability of other expertise.
Despite this lack of an objective way of indicating sufficient confidence, having a measure of it can express how much the model's decision should be trusted.
We presented and described the inter-relationship between confidence, balanced accuracy, and population, which allows for customisation of the acceptance criteria of the decision depending on the constraints.
This can be used to decide whether each individual's outcome should be trusted.
Furthermore, by combining the confidence with the ranks, the model can signal the most important features that it needs to strengthen its confidence in the output, which would accelerate additional investigations.

Not only did CACTUS prove to have better classification accuracy than traditional ML models, but it also provided insights into its reasoning. 
Thanks to its transparency, we could compare the elements the algorithm used for the classification.
These elements match what is expected from the literature.
One of the major factors in AMD is age~\cite{Salimiaghdam2020}, and CACTUS identified age as the highest rank in determining the stage of the disease.
In addition to age, the total genetic score, complement genetic score, and ARSM2 genotype were ranked highly by CACTUS in AMD classification.
CACTUS also identified the importance of a slow decline in best-corrected visual acuity as contributing to its prediction.
Visual outcomes decline slowly, approximately $1.5-2$ letters per year in AMD~\cite{Keenan2020}.
Some of the features identified as informative on the dataset exhibited an irregular trajectory across the AMD stages.
These features were dietary information.
These were under-reported in the dataset, showing a noisy and biased distribution.
Interestingly, the distributions of these features change the most between stages 3 and 4, probably because dietary interventions are suggested to slow the disease progression, and participants take these as their prognoses worsen~\cite{Salimiaghdam2020}.
Dietary behaviour can change during the disease progression~\cite{Chiu2009}, and, in this case, the proposed approach was able to capture this and exploit it to perform better classification on the few patients who have this information.

We propose CACTUS for diagnosing AMD stages in an explainable way, allowing clinicians to make quicker diagnoses and focus on patients requiring clinical attention, thus optimising scarce resources.
Since our goals include reducing the time burden on the clinicians, providing early diagnosis, and allowing for large-scale screenings, we did not train our model on features obtained by medical images; the only exception is the AMD classification that the model aims to predict, which was obtained by grading retinal images.
Retinal features are widely analysed and considered in other works on AMD classification and prediction~\cite{Yan2020, Burlina2017, Burlina2018, Schmidt-Erfurth2018-artificial, Michl2020, Perepelkina2021, Chew2020, Ting2017, Bogunovic2017}.
We still considered general ophthalmic variables, such as visual acuity, as they are easily obtainable without the mentioned constraints. 
Through the Eye-Risk consortium, we gained access to a dataset on AMD curated by the E3 consortium, on which we applied CACTUS.
The Eye-Risk and E3 consortia are two public European research entities specialised in eye diseases, to foster data collection, sharing and collaboration among numerous research groups.
This dataset harmonises multiple European studies on AMD, providing a faceted representation of its progression.
We found that CACTUS provided representations of each AMD stage and a transparent classification that could be evaluated and tuned against the literature, with high resilience to missing data.
Building a holistic representation of the interactions between each pair of elements allows for capturing the complex interdependence AMD relies on.
This approach can overcome the challenges of recognising AMD without the information provided by retinal images. 
We want to exploit the ability of CACTUS to perform classification on smaller and fragmented datasets, combined with a novel notion of confidence, to provide a useful and reliable method for explainable automatic AMD classification.
Using only the available data, CACTUS can be applied more quickly and smoothly in real scenarios, and clinicians can exploit its transparency to validate its reasoning and increase their trust in this tool.
This flexibility, uncommon in ML and DL algorithms, can allow for screening large populations, opening new possibilities in healthcare.

\section*{Methods}
CACTUS provides explainable classifications that are resilient to missing values and noise by applying abstractions.
CACTUS ingests data in a tabular format, performs a discretization of its continuous features, and considers the interactions between them to extract meaningful patterns and interactions.
We expanded the original tool~\cite{Gherardini2024} to better abstract datasets with more than two classes, compute the interactions between highly categorical features more accurately and provide a measure of how confidently each classification was performed.
Figure~\ref{fig:pipeline} depicts the general steps performed by CACTUS, which will be detailed in the following paragraphs.

Our experiment was conducted on a dataset comprising multiple European studies curated by the EYE-RISK project, through which multidimensional data collection was generated~\cite{Delcourt2016}.
It gathered genetic, life-related, medical and ophthalmic data from different Europe-wide studies to provide a comprehensive, multifaceted dataset on AMD.
We did not input features extracted from retinal images to CACTUS to prove that a meaningful screening approach of AMD is possible.
The excluded ophthalmic variables included the presence and size of drusen, Retinal Pigment Epithelium (RPE) detachments, and area and location of dry and wet AMD.
The eye with the most advanced AMD determined patients' severity and ranged from 0 (healthy) to 4 (late AMD).
Altogether, at most 5 visits per patient were recorded on average every 4 years, covering 25 years.
Each visit contains dynamic data such as age, visual acuity at a given visit, blood samples and prescribed drugs.
Immutable information, such as genomic data, was recorded only during the first visit and comprised the Single Nucleotide Polymorphisms (SNPs) of 52 common variants associated with AMD.
Due to the duration of the study, many patients did not participate in all 5 visits.
New patients were enrolled during the considered period, but the turnover was insufficient to replenish the lost participants, leading the later visits to include increasingly fewer people.
Furthermore, the proportions between AMD stages are skewed towards healthy individuals for all visits, especially the first.
To have both a populous and balanced dataset, we considered the last visit available for each person, obtaining the desired properties.
The defined dataset contains $29908$ patients and $218$ features.
The individuals are 55.46\% without AMD, 26.85\% with stage 1, 10.76\% with stage 2, 3.15\% with stage 3, and 3.78\% with stage 4.
A schematic representation of the visits is shown in Table~\ref{tab:dataset-stages}.

\subsection*{Abstractions}
CACTUS transforms continuous values into discrete elements using an abstraction process.
These discrete elements are named \emph{Up} ($\_U$) and \emph{Down} ($\_D$) \emph{flips}.
The abstraction process includes identifying an optimal \emph{cut-off value} to partition the continuous features.
To achieve this transformation, the available classes (i.e., the AMD diagnosis) are temporarily partitioned into two sets, and the values of each feature are considered as a potential discriminator between these two groups.
For each feature value, the Receiver Operating Characteristic (ROC) curve is built to consider how well it separates the two sets, and the one returning the highest Balanced Accuracy is used as a cut-off value to transform the continuous values into flips.
For instance, $60$ years of age may be the most effective cut-off for separating the first class (i.e., no AMD) from the others (i.e., any AMD form); therefore, people with at most $60$ years of age will have the flip \emph{Age\_D}, and all those who are older will have \emph{Age\_U}.
Indeed, this approach is straightforward when two classes are available, but it is less straightforward when more are available.
The division of the classes into two groups was arbitrary in previous versions of CACTUS.
This can be problematic even when there is a plausible differentiation (e.g., no AMD and any form of AMD), as not all features may discriminate between AMD stages in the same way.
Some elements could better separate a certain partitioning (e.g., no/early AMD and late AMD) than another one (e.g., no AMD and any AMD), or even just a particular AMD stage (e.g., geographic atrophy) from the rest.
For this reason, we present an extension to cover this case, which is an exhaustive search for the best cut-off value over all possible class partitioning and feature values.
This guarantees the maximum discriminatory power for each attribute while maintaining the simplicity of having a two-level abstraction.
This transformation is applied only to continuous values, which CACTUS recognises automatically by checking if a feature has at most $10$ integer values or is specified in its configuration file.
Features that are recognised as categorical are just formatted in the same way as the abstracted values (e.g. ARMS2\_rs3750846\_0, ARMS2\_rs3750846\_1, and ARMS2\_rs3750846\_2 represent the three allelic configurations of this SNP).

\subsection*{The Knowledge Graphs}
Transforming continuous values into discrete elements allows for computing their probability of being related to each AMD stage and to one another and having a more intuitive representation of reality.
CACTUS builds weighted directed graphs using the graph-tool library (https://graph-tool.skewed.de)~\cite{PEIXOTO_GRAPH-TOOL2014}.
The connections represent the conditional probability of observing a flip given another.
This way, let $A\_U$ be the \emph{Up} flip of feature A, and $B\_U$ the equivalent flip of feature B, the conditional probability of $B\_U$ given $A\_U$ in class $c$ is:

\begin{align}
    P(B\_U|A\_U, c) = \frac{P(A\_U \cap B\_U|c)}{P(A\_U|c)}.
    \label{eq:conditional-prob-pagerank}
\end{align}
This might represent, for instance, the likelihood of having high blood pressure (i.e., Blood\_pressure\_U), being old (i.e., Age\_U), and having intermediate AMD (i.e., stage 2).
The conditional probability is computed only between flips belonging to different features, as flips of the same variables cannot appear together in the same patient (e.g., an individual cannot be simultaneously young and old).
As a result, the graph for each class is not fully connected in terms of flips, but it is in terms of features.
An improvement over the previous version of CACTUS is how the conditional probabilities are used: rather than weighing the edges by the conditional probability, their absolute distance from $0.5$, which represents independent variables, is used.
This way, the connection is weighed by the strength of the influence between the flips.
Indeed, this way of connecting flips does not differentiate between positive and negative correlations, but the PageRank and Degree algorithms, explained below, require positive edges to operate correctly. 

\subsection*{Graph centrality}
Capturing and tracking the interactions of many elements in a disease is fundamental to understanding its causes, origins, and potential treatments.
Initially, CACTUS computed the centrality of each component of a particular class exclusively through the PageRank algorithm.
This procedure, originally designed and used in Google, ranks web pages using the number and importance of their web links, considering websites referenced by others as more influential.
A page referred to by well-linked pages gains a high score itself.
PageRank has been used in many scientific disciplines besides web analysis, such as biology and chemistry~\cite{GLEICH2015}.
One of the vulnerabilities met during CACTUS development~\cite{Gherardini2024} was discovered when applying PageRank on graphs with highly categorical features, as the global centrality becomes more homogeneous as the number of connections increases.
As a result, these features flatten the centrality, and the network centrality becomes less informative and decisive.
The solution we chose was the \emph{total degree} of a flip, which is the sum of the weights of the incoming and outgoing connections.
The total degree tracks how well each flip is \emph{locally} interconnected without being reduced by the number of connections in the network, which helped in datasets where categorical features with high ($>10$) unique values were stored (data not shown).

After the centrality was computed, the three classification metrics in CACTUS were computed.
These are CACTUS Probabilistic (CPB), Degree (CDG), and PageRank (CPR).
The first only considers the independent association of each flip to each class, while the other two multiply this metric by the graph centrality they represent.
This way, the CPB and the graph centrality-based algorithms, CPR and CDG, model three distinct representations of the flip \emph{significance}.

\subsection*{Classification}
Let $\sigma(c, x_i, m)$ be the significance of the flip $x_i$ of the feature $x$ in the class $c$ for the metric $m$, then the similarity to the class $c$ can be defined as $C_{c, m}=\sum\limits_{i\in N} \sigma(c, x_i, m)$ for all the $N$ features.
Therefore, every flip contributes to the cost function of each class with its corresponding significance.
The class achieving the highest score is the most similar to the sample and is assigned.
In the previous CACTUS publication, the product operation was applied over the flip significances, which represented a vulnerability in situations where flips had centralities close or equivalent to 0, which is very likely in real-case datasets with many missing values.
We used CACTUS to build a representation of each AMD stage and assign the most similar one to each individual contained in the dataset.
CPB, CPR, and CDG assign different significance scores to the features, providing three distinct classifications.

\subsection*{Interpretation}
CACTUS gives the user a detailed representation of the patterns it extracts from the dataset and a measure of confidence in each decision.
Confidence and Rank are the two metrics useful in making CACUS \emph{trustworthy} and \emph{interpretable}.

\subsubsection*{Rank}
The Rank measures how much the significance of the flips of a feature changes between classes, as this difference drives the classification towards the final result.
As previously mentioned, the relevance of a flip is obtained by the three classification metrics available in CACTUS: Probabilistic (CPB), PageRank (CPR), and Degree (CDG).
First, we compute the Rank $R_{m, x_f}$ of a single flip $x_f$ of a feature $x$ for the metric $m$ as
\begin{equation}
	R_{m, x_f} = \frac{\sum\limits_{i=1}^{N-1} \sum\limits_{j>i}^{N} |\sigma(c_i, m, x_f) - \sigma(c_j, m, x_f)|}{_NC_2},
	\label{eq:rank}
\end{equation}
which considers how the significance $\sigma(c_i, m, x_f)$ will change, on average, across the $_NC_2$ unsorted pairs of $N$ class instances.
The unordered pairs $(c_i, c_j)$ do not assume the classes to be cardinally arranged.
Intuitively, the more the significance of $x_f$ changes, the more valuable it will be for classification, as it will provide different contributions to the different classes, lowering or raising the similarity with it.
The Rank can then be averaged over the $\boldsymbol{F}_x$ flips of the feature $x$ as
\begin{equation}
	\bar{R}_{m, x} = \frac{1}{|\boldsymbol{F}_x|}\sum\limits_{f \in \boldsymbol{F}_x} R_{m, x_f}.
	\label{eq:avg-rank}
\end{equation}
The average Rank describes the influence of the feature in differentiating the available classes.

\subsubsection*{Confidence}
Determining the most powerful features in discriminating between the available classes allows for validating the model behaviour and checking for biases and/or erratic reasoning.
On the other hand, understanding how \emph{confident} the model is when it assigns a label is a valuable metric to know how much it should be \emph{trusted} when assigning an outcome.
We computed the confidence in a patient's outcome as the average absolute difference between the cost of each class $i$, $C_i$.
This is modelled as
\begin{equation}
    \mathrm{Confidence} = \frac{\sum\limits_{i \neq m}^{N} |C_m - C_i|}{N-1},
    \label{eq:confidence}
\end{equation}
where $N$ is the number of classes, and $m$ is the class that achieved the maximum similarity during classification, formally defined as $m=argmax(C)$.
Intuitively, the model is very confident if only one class has a high significance, while it is more uncertain when all significance scores are close.
The confidences are then normalised in the interval $[0, 100]$.

Guidotti et al.~\cite{Guidotti2021} defined an important pillar of explainable systems called \emph{knowledge limits}.
They indicate cases in which the model was not designed or approved for, or when the model's knowledge is insufficient to answer reliably.
Intuitively, when the model indicates low confidence, the user could provide additional information or rely more on the clinician to obtain a more solid classification.
The definition of sufficient confidence evades a strict and universal definition but is somewhat dependent on the application, such as the allowed risk, the cost of gathering more information to increase the confidence of the model or consulting other specialists~\cite{Nouretdinov2011}.

\subsection*{Comparison against standard Machine Learning algorithms}
To showcase the full potential of CACTUS in differentiating AMD, we chose standard ML models as contenders: Ridge~\cite{Johansesn1997}, Logistic Regression (LR)Z~\cite{LaValley2008}, Support Vector Machine (SVM)~\cite{Cortes1995}, Stochastic Gradient Descent (SGD)~\cite{Amari1993}, Random Forest (RF)~\cite{Breiman2001}, and XGBoost (XGB)~\cite{Chen2016}.
All were directly available in the scikit-learn Python library~\cite{python, scikit-learn}, but XGBoost was available through a dedicated Python package.
This was the first time we compared CACTUS against XGBoost, which can also natively deal with missing values.
The AMD features were scaled using the \emph{StandardScaler} class available before inputting them to ML models, and a $80/20\%$ split for the training/testing phases was applied.
A 10-fold Cross-Validation was applied to each model to provide consistent results.
For all the considered ML models but XGBoost, the missing values were filled with the average value in the corresponding column; XGBoost did not need this procedure as it automatically learns how to handle missing values~\cite{Chen2016}.

CACTUS makes the most out of small datasets due to its measures against overfitting.
As described by Ying~\cite{Ying2019}, overfitting happens due to ``the presence of noise, the limited size of the training set, and the complexity of classifiers''.
CACTUS reduces the noise contained in the dataset by applying its abstraction mechanism and exploiting a slim set of parameters, corresponding to one weight (significance) per flip for each class, to perform classification.
Finally, as described by White~\cite{White1994} and Hellstr{\"{o}}m~\cite{Hellstrom1998}, ``in the statistical context there is no such thing as overtraining when there is a weak model that has low complexity and operates on data with low-level noise''.
Since CACTUS does not make any assumptions (weak model) on how the real process works, since it builds an almost fully connected graph between flips to describe each class, and considers very few parameters, none of them set \emph{a priori}, after applying abstractions to reduce noise, we can consider CACTUS to mitigate overtraining.

We incrementally and randomly removed values from the EYE-RISK dataset to demonstrate the model's capability and practical utility in real scenarios, particularly its resilience to missing values.
Therefore, CACTUS and the other ML models were tested against the original and fragmented versions of the dataset (20\%, 40\%, 60\%, and 80\% of the values were randomly removed).
Notably, the original dataset already exhibited missing information.
After performing this comparison, we used the confidence metric to determine which of the three classification methods (CPB, CPR, and CDG) was the most reliable in real-case scenarios; this evaluation chose the model with the highest Balanced Accuracies for the 50\%, 60\%, 70\%, 80\% and 90\% of the population.
This criterion is because such a model is more trustworthy, as it will generally perform better for every confidence threshold.

\section*{Acknowledgements}
We are deeply thankful to the EYE-RISK consortium:
Soufiane Ajana\footnotemark[1],
Audrey Cougnard-Grégoire\footnotemark[1],
Cécile Delcourt\footnotemark[1],
Bénédicte M.J. Merle\footnotemark[1],
Blanca Arango-Gonzalez\footnotemark[2],
Sascha Dammeier\footnotemark[2],
Sigrid Diether\footnotemark[2],
Sabina Honisch\footnotemark[2],
Ellen Kilger\footnotemark[2],
Verena Arndt\footnotemark[3],
Tanja Endermann\footnotemark[3],
Vaibhav Bhatia\footnotemark[4],
Shomi S. Bhattacharya\footnotemark[4],
Sofia M. Calado\footnotemark[4],
Berta De la Cerda\footnotemark[4],
Francisco J. Diaz-Corrales\footnotemark[4],
Marc Biarnés\footnotemark[5],
Anna Borrell\footnotemark[5],
Lucia L. Ferraro\footnotemark[5],
Míriam Garcia\footnotemark[5],
Jordi Monés\footnotemark[5],
Eduardo Rodríguez\footnotemark[5],
Sebastian Bühren\footnotemark[6],
Johanna M. Colijn$^{7,8}$,
Magda Meester-Smoor$^{7,8}$,
Elisabeth M. van Leeuwen$^{7,8}$,
Timo Verzijden$^{7,8}$,
Caroline C.W. Klaver$^{7,8,9}$,
Eiko K. de Jong\footnotemark[9],
Thomas J. Heesterbeek\footnotemark[9],
Carel B. Hoyng\footnotemark[9],
Eveline Kersten\footnotemark[9],
Anneke I. den Hollander$^{9,10}$,
Eszter Emri\footnotemark[11],
Imre Lengyel\footnotemark[11],
Hanno Langen\footnotemark[12],
Cyrille Maugeais\footnotemark[12],
Everson Nogoceke\footnotemark[12],
Phil Luthert\footnotemark[13],
Tunde Peto\footnotemark[14],
Frances M. Pool\footnotemark[15],
Marius Ueffing$^{2,16}$,
Karl U. Ulrich Bartz-Schmidt$^{2,16}$,
and Markus Zumbansen\footnotemark[17].
And the European Eye Epidemiology Consortium for granting us access to the data used in this study:
Niyazi Acar\footnotemark[16],
Eleftherios Anastosopoulos\footnotemark[17],
Augusto Azuara-Blanco\footnotemark[18],
Arthur Bergen\footnotemark[19],
Geir Bertelsen\footnotemark[20],
Christine Binquet\footnotemark[21],
Alan Bird\footnotemark[22],
Lionel Br\'etillon\footnotemark[16],
Alain Bron\footnotemark[21],
Gabrielle Buitendijk\footnotemark[43],
Maria Luz Cachulo\footnotemark[23],
Usha Chakravarthy\footnotemark[18],
Michelle Chan\footnotemark[24],
Petrus Chang\footnotemark[25],
Johanna M. Colijn\footnotemark[7,8],
Audrey Cougnard-Gr\'egoire\footnotemark[26],
Catherine Creuzot-Garcher\footnotemark[21],
Philippa Cumberland\footnotemark[27],
Jos\'e Cunha-Vaz\footnotemark[23],
Vincent Daien\footnotemark[28],
Gabor Deak\footnotemark[29],
C\'ecile Delcourt\footnotemark[26],
Marie-Noëlle Delyfer\footnotemark[26],
Anneke den Hollander\footnotemark[30],
Martha Dietzel\footnotemark[31],
Maja Gran Erke\footnotemark[20],
Sascha Fauser\footnotemark[32],
Robert Finger\footnotemark[25],
Astrid Fletcher\footnotemark[33],
Paul Foster\footnotemark[24],
Panayiota Founti\footnotemark[17],
Arno Göbel\footnotemark[25],
Theo Gorgels\footnotemark[19],
Jakob Grauslung\footnotemark[34],
Franz Grus\footnotemark[35],
Christopher Hammond\footnotemark[36],
Catherine Helmer\footnotemark[26],
Hans-Werner Hense\footnotemark[31],
Manuel Hermann\footnotemark[32],
Ren\'e Hoehn\footnotemark[35],
Ruth Hogg\footnotemark[18],
Frank Holz\footnotemark[25],
Carel Hoyng\footnotemark[30],
Nomdo Jansonius\footnotemark[43],
Sarah Janssen\footnotemark[19],
Anthony Khawaja\footnotemark[24],
Caroline Klaver\footnotemark[43],
Jean-François Korobelnik\footnotemark[26],
Julia Lamparter\footnotemark[35],
M\'elanie Le Goff\footnotemark[26],
Sergio Leal\footnotemark[23],
Yara Lechanteur\footnotemark[30],
Terho Lehtimäki\footnotemark[37],
Andrew Lotery\footnotemark[38],
Irene Leung\footnotemark[22],
Matthias Mauschitz\footnotemark[25],
B\'en\'edicte Merle\footnotemark[26],
Verena Meyer zu Westrup\footnotemark[31],
Edoardo Midena\footnotemark[39],
Stefania Miotto\footnotemark[39],
Alireza Mishahi\footnotemark[35],
Sadek Mohan-Saïd\footnotemark[40],
Alyson Muldrew\footnotemark[18],
Michael Mueller\footnotemark[37],
Sandrina Nunes\footnotemark[23],
Konrad Oexle\footnotemark[41],
Tunde Peto\footnotemark[22],
Stefano Piermarocchi\footnotemark[39],
Elena Prokofyeva\footnotemark[28],
Jugnoo Rahi\footnotemark[24],
Olli Raitakari\footnotemark[37],
Luisa Ribeiro\footnotemark[23],
Maria-B\'en\'edicte Rougier\footnotemark[26],
Jos\'e Sahel\footnotemark[40],
Aggeliki Salonikiou\footnotemark[17],
Clarisa Sanchez\footnotemark[30],
Steffen Schmitz-Valckenberg\footnotemark[25],
C\'edric Schweitzer\footnotemark[26],
Tatiana Segato\footnotemark[39],
Jasmin Shehata\footnotemark[29],
Rufino Silva\footnotemark[23],
Giuliana Silvestri\footnotemark[18],
Christian Simader\footnotemark[29],
Eric Souied\footnotemark[42],
Henriet Springelkamp\footnotemark[43],
Robyn Tapp\footnotemark[37],
Fotis Topouzis\footnotemark[17],
Virginie Verhoeven\footnotemark[43],
Therese Von Hanno\footnotemark[20],
Stela Vujosevic\footnotemark[39],
Katie Williams\footnotemark[36],
Christian Wolfram\footnotemark[35],
Jennifer Yip\footnotemark[24],
Jennyfer Zerbib\footnotemark[42],
and Isabella Zwiener\footnotemark[35].

\renewcommand{\footnoterule}{%
  \kern -3pt
  \hrule width \textwidth height 1pt
  \kern 2pt
}
\renewcommand{\footnotesize}{\scriptsize}

\vspace*{\fill} 
\begin{minipage}{\textwidth}
\begin{multicols}{3}
\footnotesize
\begin{enumerate}
\setcounter{enumi}{0}
\item University Bordeaux, Inserm, Bordeaux Population Health Research Center, Team LEHA, UMR 1219, Bordeaux, France.
\item Centre for Ophthalmology, Institute for Ophthalmic Research, Eberhard Karls University Tübingen, University Clinic Tübingen, Tübingen, Germany.
\item Assay Development, AYOXXA Biosystems GmbH, Cologne, Germany.
\item Department of Regeneration and Cell Therapy, Andalusian Molecular Biology and Regenerative Medicine Centre (CABIMER), Seville, Spain.
\item Barcelona Macula Foundation, Barcelona, Spain.
\item Business Development, AYOXXA Biosystems GmbH, Cologne, Germany.
\item Department of Epidemiology, Erasmus Medical Center, Rotterdam, Netherlands.
\item Department of Ophthalmology, Erasmus Medical Center, Rotterdam, Netherlands.
\item Department of Ophthalmology, Radboud University Medical Center, Nijmegen, Netherlands.
\item Department of Human Genetics, Radboud University Medical Center, Nijmegen, Netherlands.
\item Centre for Experimental Medicine, Queen's University Belfast, Belfast, United Kingdom.
\item Roche Innovation Center Basel, F. Hoffmann-La Roche Ltd, Basel, Switzerland.
\item Institute of Ophthalmology, University College London, London, United Kingdom.
\item Centre for Public Health, Queen's University Belfast, Belfast, United Kingdom.
\item Ocular Biology, UCL Institute of Opthalmology, London, United Kingdom.
\item Inra-University of Burgundy, Dijon, France.
\item University of Thessaloniki, Thessaloniki, Greece.
\item Queen's University Belfast, Belfast, the UK.
\item Netherlands Institute for Neurosciences-KNAW, Amsterdam, the Netherlands.
\item University of Tromso, Tromso, Norway.
\item University Hospital of Dijon, Dijon, France.
\item Moorfield's Eye Hospital, London, the UK.
\item AIBILI/CHUC, Coimbra, Portugal.
\item UCL Institute of Ophthalmology, London, the UK.
\item University of Bonn, Bonn, Germany.
\item University of Bordeaux Segalen, Bordeaux, France.
\item UCL Institute of Child Health, London, the UK.
\item Inserm U1061, Montpellier, France.
\item Medical University of Vienna, Vienna, Austria.
\item Radboud University, Nijmegen, the Netherlands.
\item University of Muenster, Muenster, Germany.
\item University Eye Hospital, Cologne, Germany.
\item London School of Hygiene and Tropical Medicine, London, the UK.
\item University of Southern Denmark, Odense, Denmark.
\item University Medical Center Mainz, Mainz, Germany.
\item King's College, London, the UK.
\item Pirkanmaa Hospital District, Tampere, Finland.
\item University of Southampton, Southampton, the UK.
\item University of Padova, Padova, Italy.
\item Institut de la Vision, Paris, France.
\item Institute of Human Genetics, Munich, Germany.
\item University Hospital of Cr\'eteil, Cr\'eteil, France.
\item Erasmus Medical Center, Rotterdam, the Netherlands.
\end{enumerate}
\end{multicols}
\end{minipage}
\clearpage

This publication is supported by two projects funded by the European Union’s Horizon 2020 Research and Innovation Programme: the EYE-RISK project - grant agreement No 634479, and the Sano project - grant agreement No 857533. 
The Sano Project has also received funding from the International Research Agendas Programme of the Foundation for Polish Science - No MAB PLUS/2019/13, and the Minister of Science and Higher Education ``Support for the activity of Centres of Excellence established in Poland under Horizon 2020'' based on the contract number MEiN/2023/DIR/3796.
Luca Gherardini has been employed at the Sano Centre for Computational Medicine and supported by the Queen’s University Belfast Postgraduate Support.
The Kelvin-2 research cluster of the NI-HPC centre at Queen’s University Belfast supported this research.

The authors are deeply grateful to Professor Roger Woods and Varun Ravi Varma, who provided invaluable suggestions for improving the quality of this manuscript.

\section*{Authors' contributions}
Conceptualisation: LG, IL, JS.
Data curation: IL, TP, CCWK, MAMS, JMC.
Formal analysis: LG, IL, TP.
Investigation: LG, IL.
Methodology: LG, JS.
Software: LG.
Supervision: JS, IL, TP.
Validation: JS, IL, TP.
Visualisation: LG.
Writing - original draft: LG, IL, TP.
Writing - review and editing: LG, IL, TP, CCWK, MAMS, JMC, JS.

\section*{Declaration of interests}
The authors do not have any conflicting interests to declare.

\section*{Ethics statement}
This study complied with relevant laws and institutional guidelines. 
The data processed in this study were collected before this work.
Therefore, no ethical approval was required.
The personal information of each individual was anonymised. 

\section*{Figures, figure titles, and figure legends}

\clearpage
\section*{Tables, table titles, and table legends}

\begin{table}[h!]
\resizebox{\textwidth}{!}{%
\begin{tabular}{ccccccccc}
\multicolumn{1}{l}{} & \multicolumn{1}{l}{} & \multicolumn{7}{c}{Visit} \\
 & \multicolumn{1}{c|}{} & 1 & 2 & 3 & 4 & \multicolumn{1}{c|}{5} & First & Last \\ \hline
\multirow{5}{*}{AMD Stage} & \multicolumn{1}{c|}{0} & 18168 (62.84\%) & 6590 (55.25\%) & 2588 (40.52\%) & 1239 (39.62\%) & \multicolumn{1}{c|}{528 (34.33\%)} & 18624 (62.27\%) & 16588 (55.46\%) \\
 & \multicolumn{1}{c|}{1} & 6790 (23.49\%) & 3911 (32.79\%) & 2599 (40.69\%) & 1290 (41.25\%) & \multicolumn{1}{c|}{580 (37.71\%)} & 7161 (23.94\%) & 8029 (26.85\%) \\
 & \multicolumn{1}{c|}{2} & 2497 (8.64\%) & 1012 (8.48\%) & 696 (10.9\%) & 377 (12.06\%) & \multicolumn{1}{c|}{255 (16.58\%)} & 2616 (8.75\%) & 3219 (10.76\%) \\
 & \multicolumn{1}{c|}{3} & 636 (2.2\%) & 224 (1.88\%) & 216 (3.38\%) & 112 (3.58\%) & \multicolumn{1}{c|}{116 (7.54\%)} & 657 (2.2\%) & 942 (3.15\%) \\
 & \multicolumn{1}{c|}{4} & 819 (2.83\%) & 190 (1.59\%) & 288 (4.51\%) & 109 (3.49\%) & \multicolumn{1}{c|}{59 (3.84\%)} & 850 (2.84\%) & 1130 (3.78\%) \\ \hline
 & \multicolumn{1}{c|}{Total} & 28910 & 11927 & 6387 & 3127 & \multicolumn{1}{c|}{1538} & 29908 & 29908 \\
 & \multicolumn{1}{c|}{Dropped} &  & 17526 & 6775 & 3560 & \multicolumn{1}{c|}{1695} &  &  \\
 & \multicolumn{1}{c|}{New} &  & 543 & 1235 & 300 & \multicolumn{1}{c|}{106} &  &  \\ \hline
\multirow{5}{*}{Average age per stage} & \multicolumn{1}{c|}{0} & 66.76 & 66.23 & 71.65 & 74.24 & \multicolumn{1}{c|}{78.3} & 66.87 & 69.56 \\
 & \multicolumn{1}{c|}{1} & 69.61 & 68.94 & 72.77 & 74.51 & \multicolumn{1}{c|}{78.31} & 69.83 & 73.0 \\
 & \multicolumn{1}{c|}{2} & 72.6 & 71.86 & 75.21 & 76.46 & \multicolumn{1}{c|}{79.86} & 72.75 & 74.6 \\
 & \multicolumn{1}{c|}{3} & 74.56 & 74.72 & 76.83 & 78.97 & \multicolumn{1}{c|}{80.34} & 74.56 & 76.78 \\
 & \multicolumn{1}{c|}{4} & 76.32 & 80.06 & 79.3 & 81.24 & \multicolumn{1}{c|}{83.53} & 76.61 & 78.44
\end{tabular}%
}
\caption{
\textbf{Description of each visit in the EYE-RISK dataset}. Additional groups comprising, respectively, the first and last visit of each patient have been considered.
For each group, the number and percentage of people affected by a particular AMD stage have been reported, in addition to the number of dropped and newly enrolled patients.
The average age (in years) was reported during each visit.
It is noticeable how healthy patients (first row) are the most prevalent stage recorded before declining towards the end of the study.
While this ensures a more balanced dataset, the total amount of patients is also dramatically decreasing due to heavy dropouts.
Considering the First and, in particular, the Last visit for each patient, we can obtain sensibly larger cohorts and more balanced proportions.
Due to the better balance of the Last visit, we selected this group for further analysis.
}
\label{tab:dataset-stages}
\end{table}

\begin{table}[h!]
\resizebox{\textwidth}{!}{%
\begin{tabular}{c|ccc|cccccc}
Missing values added to Last Visit & CDG & CPB & CPR & LR & RF & Ridge & SGD & SVM & XGB \\ \hline
0\% & 0.33±0.01 & 0.32±0.01 & \textbf{0.34±0.01} & 0.31±0.01 & 0.3±0.01 & 0.29±0.01 & 0.31±0.01 & 0.27±0.0 & 0.31±0.01 \\
20\% & 0.31±0.01 & 0.32±0.02 & \textbf{0.33±0.01} & 0.29±0.01 & 0.28±0.0 & 0.27±0.0 & 0.29±0.01 & 0.26±0.0 & 0.3±0.01 \\
40\% & \textbf{0.3±0.01} & \textbf{0.3±0.01} & \textbf{0.3±0.01} & 0.28±0.01 & 0.27±0.0 & 0.25±0.01 & 0.28±0.01 & 0.25±0.0 & 0.29±0.0 \\
60\% & \textbf{0.3±0.02} & \textbf{0.3±0.02} & 0.29±0.02 & 0.26±0.0 & 0.26±0.0 & 0.23±0.0 & 0.26±0.01 & 0.23±0.01 & 0.28±0.0 \\
80\% & \textbf{0.28±0.01} & \textbf{0.28±0.01} & 0.27±0.01 & 0.24±0.0 & 0.23±0.0 & 0.21±0.0 & 0.24±0.01 & 0.21±0.0 & 0.25±0.0
\end{tabular}%
}
\caption{
\textbf{The balanced accuracy, in percentage, achieved by CACTUS's modalities \emph{Degree} (CDG), \emph{Probabilistic} (CPB), and \emph{PageRank} (CPR) in comparison to standard ML techniques}.
In addition to the original dataset of the last visits, we extended the analysis to different percentages of induced missing values to test the model's resilience.
All CACTUS modalities performed better than ML models for all datasets, with PageRank achieving the highest performance in 3 out of 5 experiments.
All models have been tested through a 10-fold cross-validation approach and an 80/20 split for train/test sets. 
The standard deviation across the cross-validation was included along the average balanced accuracy.
}
\label{tab:classification-results}
\end{table}

\begin{table}[]
	\resizebox{\textwidth}{!}{%
		\begin{tabular}{c|ccc}
			& BA on the whole dataset (\%)& BA on the refined dataset (\%)& BA on the 9 highest ranks (\%)\\ \hline
			CDG & 33±0.01& 26±0.00& 31±0.01\\
			CPB & 32±0.01& 33±0.01& 29±0.01\\
			CPR & 34±0.01& 34±0.01& 29±0.01\end{tabular}%
	}
\caption{
\textbf{
The Balanced Accuracy (BA) achieved by the three classification metrics in CACTUS for the whole dataset, a refined set of features, and a selection of the 9 most important features}. 
The dataset was refined by excluding features without biological significance for AMD or with more than $50\%$ missing values. 
The 9 features with the highest rank in this refined version are shown in Figure~\ref{fig:corrected-ranks}. 
The features in these three sets were $173$, $112$, and $9$.
}
\label{tab:ba-comparison}
\end{table}

\clearpage

\bibliographystyle{elsarticle-num}
\bibliography{bibliography.bib}

\clearpage

\section*{Supplementary}

\begin{figure}[h!]
	\centering
	\includegraphics[width=.8\textwidth, height=\textheight, keepaspectratio]{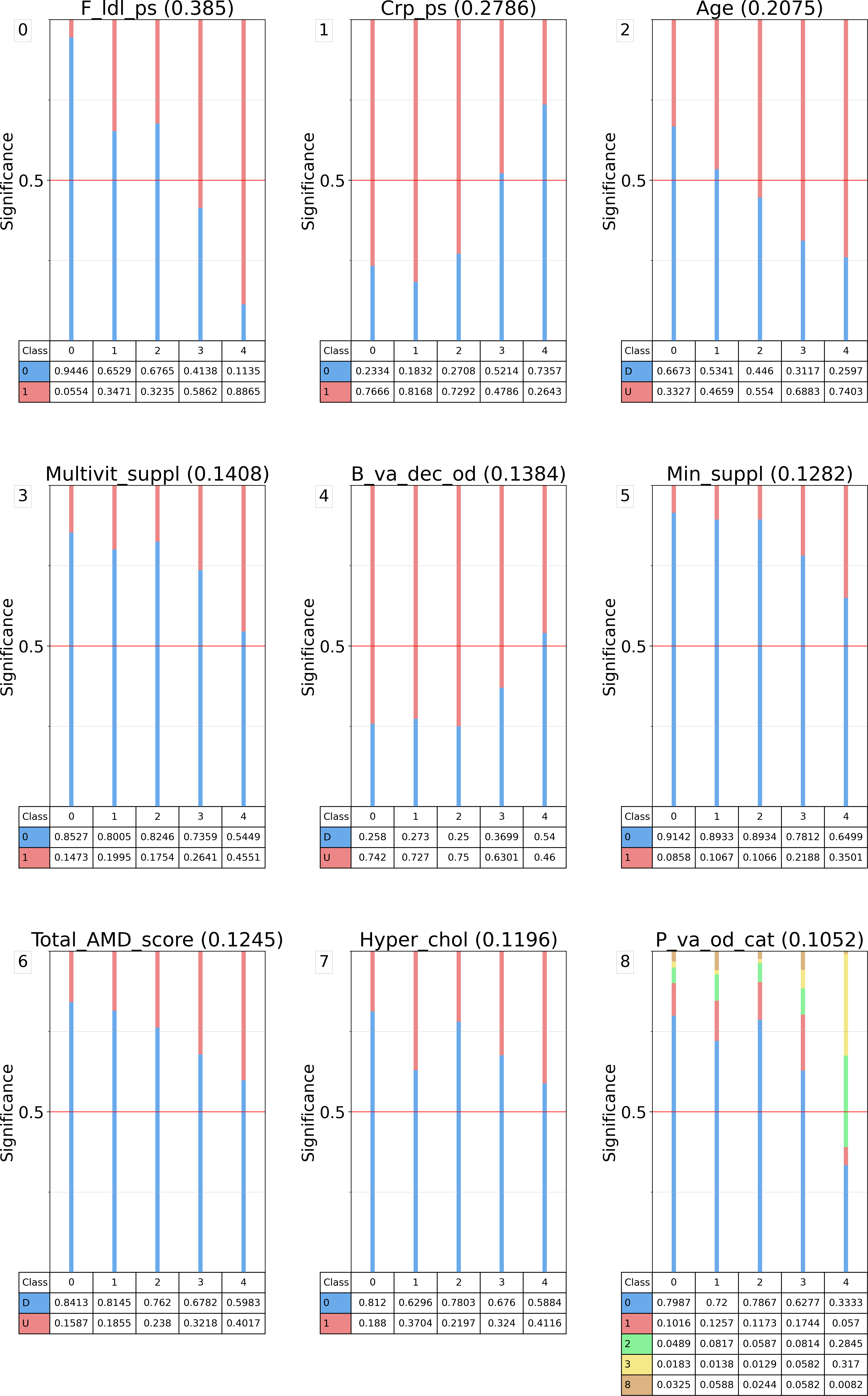}
	\caption{
		Feature ranks computed by CACTUS through the \emph{Degree} centrality on the knowledge graphs.
		They are sorted from the highest (upper left) to the lowest (lower right), as indicated in the title of each plot between parentheses.
	}
	\label{fig:ranks-degree}
\end{figure}

\begin{figure}[h!]
	\centering
	\includegraphics[width=.8\textwidth, height=\textheight, keepaspectratio]{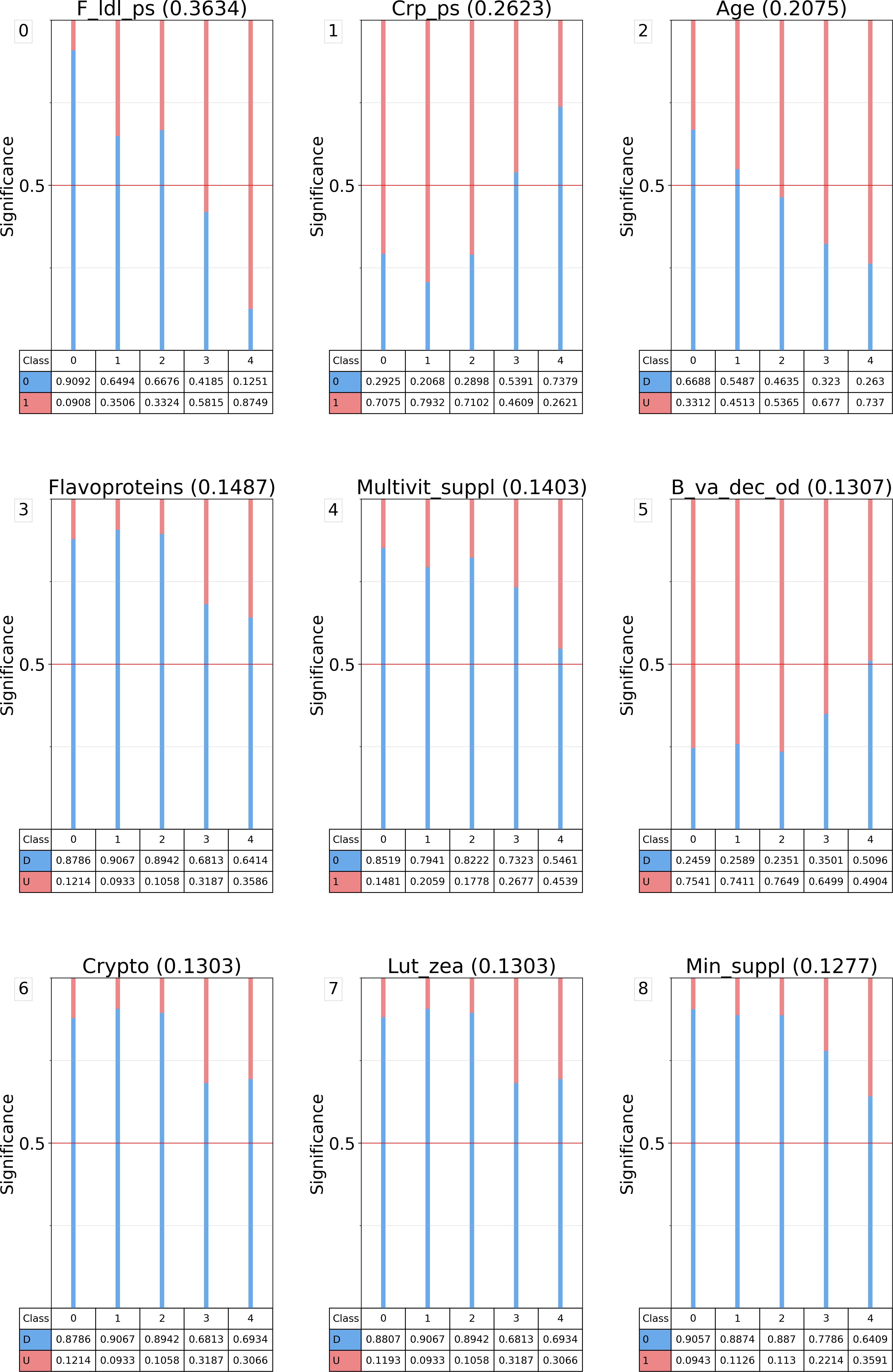}
	\caption{
		Feature ranks computed by CACTUS through the \emph{Probabilistic} significance metric.
		They are sorted from the highest (upper left) to the lowest (lower right), as indicated in the title of each plot between parentheses. The scores do not sum up to 1 because of missing values.
	}
	\label{fig:ranks-probabilistic}
\end{figure}

\LTcapwidth=\textwidth
\begin{longtable}[h!]{p{.25\textwidth}p{.05\textwidth}|p{.25\textwidth}p{.05\textwidth}|p{.25\textwidth}p{.05\textwidth}}
\multicolumn{2}{c|}{CDG} & \multicolumn{2}{c|}{CPR} & \multicolumn{2}{c}{CPB} \\
Marker & Rank & Marker & Rank & Marker & Rank \\ \hline
F\_ldl\_ps & .3963 & F\_ldl\_ps & .3845 & F\_ldl\_ps & .3748 \\
Crp\_ps & .2826 & Crp\_ps & .2609 & Crp\_ps & .266 \\
Age & .2024 & Age & .2087 & Age & .2008 \\
Multivit\_suppl & .1451 & B\_va\_dec\_od & .1415 & Acar & .1767 \\
B\_va\_dec\_od & .143 & Multivit\_suppl & .1348 & Crypto & .149 \\
Acar & .1364 & Min\_suppl & .1344 & Flavoproteins & .149 \\
Min\_suppl & .1254 & Hyper\_chol & .1198 & Lycop & .149 \\
Hyper\_chol & .1241 & Acar & .1178 & Lut\_zea & .1485 \\
Total\_AMD\_score & .1156 & Total\_AMD\_score & .1163 & Multivit\_suppl & .145 \\
P\_va\_od\_cat & .1024 & Trombo\_inh & .1103 & B\_va\_dec\_od & .1363 \\
Trombo\_inh & .1011 & B\_va\_dec\_os & .1057 & Min\_suppl & .1266 \\
ARMS2\_rs3750846 & .0953 & P\_va\_od\_cat & .1002 & Hyper\_chol & .1224 \\
B\_va\_dec\_os & .0949 & Nsaid & .0989 & Pufa\_omg6 & .1224 \\
Ses & .0921 & Ses & .098 & Total\_AMD\_score & .1116 \\
P\_va\_os\_cat & .0898 & B\_va\_os\_cat & .095 & B9 & .1049 \\
Nsaid & .0895 & B\_va\_od\_cat & .0927 & B3 & .1043 \\
CFH\_rs10922109 & .0893 & P\_va\_os\_cat & .0882 & Epa\_dha & .1036 \\
B\_va\_os\_cat & .0877 & Outside & .0844 & B1 & .1035 \\
Outside & .0867 & ARMS2\_rs3750846 & .0843 & B2 & .1035 \\
Complement\_score & .0865 & Complement\_score & .0804 & Bcar & .1026 \\
B\_va\_od\_cat & .0856 & Risk\_ARMS2 & .0771 & Pufa\_omg3 & .1026 \\
Risk\_ARMS2 & .0805 & Vitamind\_ps & .0752 & Trombo\_inh & .1016 \\
Vitamind\_ps & .077 & Fishoil\_suppl & .0702 & B\_carotene & .1012 \\
CFH\_rs570618 & .0764 & Omeg3\_suppl & .0702 & P\_va\_od\_cat & .098 \\
Fishoil\_suppl & .0752 & Iris\_color & .0672 & B\_va\_dec\_os & .0952 \\
Omeg3\_suppl & .0752 & CFH\_rs10922109 & .0662 & ARMS2\_rs3750846 & .0911 \\
Pufa\_omg6 & .074 & CFH\_rs570618 & .063 & Outside & .0874 \\
Iris\_color & .0675 & Pufa\_omg6 & .0609 & CFH\_rs10922109 & .0865 \\
Tei & .0632 & Refr\_cyl\_os & .0596 & Dairy\_freq & .0862 \\
Vitd & .0569 & F\_hdl\_ps & .0595 & Egg\_freq & .0862 \\
Vitc & .0551 & Hypertension & .0595 & Legume\_freq & .0862 \\
Vitamind & .0544 & Refr\_cyl\_od & .0585 & Vegetable\_freq & .0862 \\
Hypertension & .0534 & F\_choles\_ps & .0565 & White\_meat\_freq & .0862 \\
Cessation\_a & .053 & Vitamind & .0561 & P\_va\_os\_cat & .0854 \\
Refr\_cyl\_od & .0522 & Anti\_vegf\_any & .0551 & B\_va\_os\_cat & .0843 \\
Pressure\_sys & .0509 & Pressure\_sys & .0543 & Nsaid & .0839 \\
P\_va\_dec\_os & .0491 & Cessation\_a & .0534 & Fish\_freq & .0834 \\
Refr\_cyl\_os & .0488 & Refr\_sph\_od & .0534 & Meat\_freq & .0834 \\
Anti\_vegf\_any & .0486 & F\_trigly\_ps & .053 & Red\_meat\_freq & .0834 \\
Smoke\_ever & .0475 & Cylinder\_sign\_os & .0523 & Complement\_score & .0827 \\
Alcohol\_ever & .0473 & Tei & .052 & B\_va\_od\_cat & .0822 \\
C2\_rs114254831 & .046 & C2\_rs114254831 & .0513 & Risk\_ARMS2 & .0772 \\
F\_hdl\_ps & .0455 & P\_va\_dec\_os & .0513 & CFH\_rs570618 & .0741 \\
Refr\_axis\_od & .0454 & Alcohol\_ever & .0507 & Vitamind\_ps & .0718 \\
Refr\_sph\_od & .0448 & Cylinder\_sign\_od & .0505 & Lung\_disease & .0692 \\
Cylinder\_sign\_os & .0439 & Vitd & .0503 & Tot\_prot & .0684 \\
F\_choles\_ps & .043 & Vitc & .0481 & Iris\_color & .066 \\
Height & .0423 & Lens\_od & .048 & Ses & .0653 \\
Lens\_od & .0423 & Anti\_diab & .0477 & Hypertension & .0599 \\
Lipid\_score & .0418 & Lens\_os & .0477 & Tei & .0586 \\
Cylinder\_sign\_od & .0417 & Physical\_act & .0466 & Cessation\_a & .0566 \\
C3\_rs2230199 & .0416 & Smoke\_ever & .0462 & Zeaxanthin & .0538 \\
Anti\_hyp & .0412 & F\_glucose\_ps & .0451 & Smoke\_ever & .0529 \\
Refr\_axis\_os & .04 & P\_va\_dec\_od & .0442 & Cylinder\_sign\_os & .0518 \\
NPLOC4\_rs6565597 & .0399 & F\_trigly & .044 & Lutein & .0514 \\
F\_trigly\_ps & .0397 & Cardiovasc & .0433 & Tot\_carb & .0512 \\
ECM\_Score & .0396 & C3\_rs2230199 & .0429 & Refr\_cyl\_od & .0508 \\
Lens\_os & .0393 & NPLOC4\_rs6565597 & .0419 & Pressure\_sys & .05 \\
CFH\_rs61818925 & .0391 & Pufa & .0409 & Cylinder\_sign\_od & .0494 \\
TRPM3\_rs71507014 & .0389 & Anti\_vegf\_od & .0396 & Anti\_vegf\_any & .0492 \\
Anti\_diab & .0377 & Refr\_axis\_od & .0385 & Vitc & .0487 \\
P\_va\_dec\_od & .0368 & Anti\_glauc & .038 & F\_hdl\_ps & .0486 \\
Gfr & .0367 & Anti\_hyp & .038 & Refr\_cyl\_os & .0475 \\
LIPC\_rs2043085 & .0367 & Mufa\_sfa & .0379 & Alcohol\_ever & .0465 \\
F\_glucose\_ps & .0362 & TRPM3\_rs71507014 & .0375 & F\_choles\_ps & .0463 \\
Refr\_sph\_os & .0356 & Gfr & .0372 & Nf\_glucose & .046 \\
F\_trigly & .035 & Height & .0362 & Refr\_sph\_od & .046 \\
Cardiovasc & .0346 & Olive\_qtt & .0362 & Vitamind & .0458 \\
Anti\_vegf\_od & .0345 & Crp & .0361 & C2\_rs114254831 & .0456 \\
F\_hdl & .0339 & Lipid\_score & .0361 & Anti\_hyp & .0436 \\
CNN2\_rs67538026 & .0334 & A\_length\_od & .0359 & Vitd & .0435 \\
Crp & .0323 & Sfa & .0353 & Refr\_axis\_od & .0434 \\
Physical\_act & .0323 & Refr\_sph\_os & .0351 & Anti\_diab & .0432 \\
Alat & .032 & CNN2\_rs67538026 & .0348 & CNN2\_rs67538026 & .0425 \\
Mufa\_sfa & .0318 & Vegetable\_qtt & .0347 & F\_trigly\_ps & .0425 \\
C2\_rs943080 & .0315 & ECM\_Score & .0336 & F\_glucose\_ps & .0422 \\
Pufa & .0303 & Red\_meat\_qtt & .0327 & B12 & .0418 \\
Anti\_glauc & .0298 & Refr\_axis\_os & .0321 & B6 & .0412 \\
P\_va\_letters\_od & .0293 & P\_va\_letters\_od & .0319 & Crp & .0409 \\
Tot\_carb & .0285 & Beer & .0306 & Vite & .0409 \\
B\_va\_od & .0278 & Alzheimer & .0305 & Lens\_od & .0401 \\
Olive\_qtt & .0274 & Gamma\_gt & .03 & P\_va\_dec\_os & .04 \\
Red\_meat\_qtt & .0274 & LIPC\_rs2043085 & .03 & Refr\_axis\_os & .0399 \\
SLC16A8\_rs8135665 & .0268 & Alat & .0299 & Height & .0397 \\
Anti\_vegf\_os & .0265 & Anti\_vegf\_os & .0298 & C3\_rs2230199 & .0395 \\
A\_length\_od & .0258 & Nuts\_qtt & .029 & Lipid\_score & .0394 \\
COL8A1\_rs55975637 & .0256 & F\_hdl & .0285 & CFH\_rs61818925 & .0386 \\
Alcohol & .0254 & CFH\_rs61818925 & .0281 & Fishoil\_suppl & .0384 \\
Vegetable\_qtt & .0252 & C2\_rs943080 & .0278 & Omeg3\_suppl & .0384 \\
Alzheimer & .0251 & B\_va\_od & .0265 & NPLOC4\_rs6565597 & .0381 \\
Beer & .0243 & Wine & .0261 & Zinc\_food & .0377 \\
CETP\_rs5817082 & .0241 & SLC16A8\_rs8135665 & .0257 & F\_trigly & .0375 \\
Sfa & .0241 & Asat & .0254 & Lens\_os & .0371 \\
Bmi & .024 & Egg\_qtt & .0252 & ECM\_Score & .037 \\
Urea & .0239 & COL8A1\_rs55975637 & .0247 & Refr\_sph\_os & .037 \\
Meat\_freq & .0235 & CETP\_rs5817082 & .0244 & Anti\_vegf\_od & .0364 \\
Red\_meat\_freq & .0235 & Lipid\_low & .0237 & LIPC\_rs2043085 & .0362 \\
Vite & .0233 & Cessation\_y & .0234 & Cardiovasc & .0358 \\
PILRB\_rs7803454 & .0229 & Alcohol & .0231 & P\_va\_letters\_od & .035 \\
B6 & .0226 & A\_length\_os & .0228 & Alat & .0342 \\
Cessation\_y & .0225 & Tot\_carb & .0227 & Gfr & .0339 \\
Pressure\_od & .0225 & Bmi & .0214 & TRPM3\_rs71507014 & .0327 \\
Wine & .0222 & Oily\_qtt & .0207 & F\_hdl & .0316 \\
CETP\_rs17231506 & .0218 & Dairy\_qtt & .0206 & Urea & .0315 \\
ABCA1\_rs2740488 & .0211 & Diabetes & .0202 & C2\_rs943080 & .0312 \\
B\_va\_os & .0211 & Zeaxanthin & .0202 & Anti\_glauc & .0293 \\
Dairy\_freq & .0207 & B\_va\_os & .0201 & Physical\_act & .0293 \\
Egg\_freq & .0207 & Liquor & .0201 & Wine & .0293 \\
Legume\_freq & .0207 & MMP9\_rs142450006 & .02 & B\_va\_od & .0291 \\
Vegetable\_freq & .0207 & PILRB\_rs7803454 & .0199 & Anti\_vegf\_os & .0279 \\
Nuts\_qtt & .0202 & Pmeat\_qtt & .0199 & F\_ldl & .0279 \\
Oily\_qtt & .0196 & LIPC\_rs2070895 & .0197 & Pressure\_od & .0272 \\
RAD51B\_rs61985136 & .0195 & SYN3\_rs5754227 & .0193 & Bmi & .0264 \\
Zeaxanthin & .0193 & COL4A3\_rs11884770 & .0192 & Pack\_years & .0263 \\
Egg\_qtt & .019 & Meat\_qtt & .0188 & P\_va\_letters\_os & .0258 \\
ARHGAP21\_rs12357257 & .0189 & Urea & .0184 & Pmeat\_qtt & .0257 \\
Liquor & .0189 & B6 & .0183 & Red\_meat\_qtt & .0257 \\
LIPC\_rs2070895 & .0182 & Vite & .0183 & Alzheimer & .0256 \\
Lipid\_low & .0182 & Meat\_freq & .018 & A\_length\_od & .0253 \\
F\_choles & .018 & Red\_meat\_freq & .018 & Alcohol & .0252 \\
COL4A3\_rs11884770 & .0177 & RAD51B\_rs61985136 & .0178 & P\_va\_dec\_od & .0252 \\
Gamma\_gt & .0175 & APOE\_rs429358 & .0176 & A\_length\_os & .0245 \\
Pmeat\_qtt & .0175 & Legume\_qtt & .0176 & CETP\_rs5817082 & .0243 \\
APOE\_rs73036519 & .0171 & ABCA1\_rs2740488 & .0175 & SLC16A8\_rs8135665 & .0242 \\
TMEM97\_rs11080055 & .0166 & Fish\_qtt & .0174 & Pufa & .024 \\
Legume\_qtt & .0163 & Pressure\_od & .0171 & COL8A1\_rs55975637 & .0235 \\
Fish\_qtt & .016 & TGFBR1\_rs1626340 & .0166 & Liquor & .023 \\
CFI\_rs10033900 & .0159 & Weight & .0163 & Beer & .0228 \\
Smoke\_start & .0159 & ARHGAP21\_rs12357257 & .0158 & Mufa\_sfa & .0215 \\
TNFRSF10A\_rs79037040 & .0159 & Mufa & .0155 & Cessation\_y & .0212 \\
APOE\_rs429358 & .0158 & Dairy\_freq & .0153 & Sfa & .0206 \\
Dairy\_qtt & .0156 & Egg\_freq & .0153 & PILRB\_rs7803454 & .0205 \\
MIR6130\_rs10781182 & .0155 & Legume\_freq & .0153 & Meat\_qtt & .02 \\
W\_circ & .0155 & Vegetable\_freq & .0153 & Vegetable\_qtt & .02 \\
Asat & .0153 & F\_ldl & .0146 & ABCA1\_rs2740488 & .0199 \\
Diabetes & .015 & H\_circ & .0146 & CETP\_rs17231506 & .0199 \\
ADAMTS9\_rs62247658 & .0149 & P\_va\_letters\_os & .0145 & LIPC\_rs2070895 & .0199 \\
RAD51B\_rs2842339 & .0149 & C20orf85\_rs201459901 & .0144 & B\_va\_os & .0197 \\
SYN3\_rs5754227 & .0148 & CETP\_rs17231506 & .0143 & RAD51B\_rs61985136 & .0196 \\
Pressure\_os & .0146 & APOE\_rs73036519 & .0141 & Lipid\_low & .0191 \\
A\_length\_os & .0141 & Cigarette & .0141 & White\_qtt & .0191 \\
Meat\_qtt & .0141 & RAD51B\_rs2842339 & .0138 & ARHGAP21\_rs12357257 & .0181 \\
B3GALTL\_rs9564692 & .0138 & TMEM97\_rs11080055 & .0138 & F\_choles & .0181 \\
Tot\_prot & .0134 & B3GALTL\_rs9564692 & .0134 & Olive\_qtt & .018 \\
KMT2E\_rs1142 & .0129 & Smoke\_start & .0133 & COL4A3\_rs11884770 & .0176 \\
Lung\_disease & .0129 & MIR6130\_rs10781182 & .0127 & Smoke\_start & .0176 \\
MMP9\_rs142450006 & .0127 & Cortico & .0124 & Cereal\_qtt & .0173 \\
F\_ldl & .0126 & Pressure\_os & .0123 & APOE\_rs73036519 & .0164 \\
B12 & .0125 & Lutein & .0122 & Pressure\_os & .0164 \\
Weight & .0121 & TNFRSF10A\_rs79037040 & .0122 & MMP9\_rs142450006 & .0162 \\
Lutein & .012 & C3\_rs12019136 & .012 & SYN3\_rs5754227 & .0162 \\
TGFBR1\_rs1626340 & .0119 & ADAMTS9\_rs62247658 & .0119 & TNFRSF10A\_rs79037040 & .0155 \\
Cigarette & .0117 & F\_choles & .0119 & Fruit\_qtt & .0153 \\
Zinc\_food & .0117 & CFI\_rs10033900 & .0118 & TMEM97\_rs11080055 & .0152 \\
Fish\_freq & .0116 & CTRB2\_rs72802342 & .0115 & Diabetes & .0151 \\
C20orf85\_rs201459901 & .0113 & C2\_rs429608 & .011 & White\_meat\_qtt & .0151 \\
P\_va\_letters\_os & .0113 & Suppl\_amd & .011 & APOE\_rs429358 & .015 \\
F\_glucose & .0111 & W\_circ & .0105 & CFI\_rs10033900 & .015 \\
Cortico & .0107 & CFH\_rs187328863 & .0104 & MIR6130\_rs10781182 & .0149 \\
H\_circ & .0107 & RDH5\_rs3138141 & .0104 & Mufa & .0149 \\
C3\_rs12019136 & .0102 & C3\_rs147859257 & .0099 & Nuts\_qtt & .0147 \\
Suppl\_amd & .0094 & Fish\_freq & .0099 & RAD51B\_rs2842339 & .0144 \\
CTRB2\_rs72802342 & .0092 & KMT2E\_rs1142 & .0099 & ADAMTS9\_rs62247658 & .0143 \\
C3\_rs147859257 & .0091 & Pressure\_dias & .0098 & Gamma\_gt & .0142 \\
Pack\_years & .0091 & Pack\_years & .0096 & Weight & .0138 \\
C9\_rs62358361 & .0089 & C9\_rs62358361 & .0095 & Asat & .0137 \\
B2 & .0087 & B12 & .0094 & TGFBR1\_rs1626340 & .0131 \\
CFH\_rs187328863 & .0085 & F\_glucose & .0094 & C20orf85\_rs201459901 & .0129 \\
RDH5\_rs3138141 & .0085 & Lung\_disease & .0093 & B3GALTL\_rs9564692 & .0127 \\
Fruit\_qtt & .0083 & Fruit\_qtt & .0085 & Cigarette & .0127 \\
Mufa & .0082 & Tot\_prot & .0085 & W\_circ & .0126 \\
B1 & .0078 & Pipe & .0083 & KMT2E\_rs1142 & .0123 \\
C2\_rs429608 & .0073 & White\_meat\_qtt & .0082 & C3\_rs12019136 & .0115 \\
Pressure\_dias & .007 & Zinc\_food & .0079 & Fish\_qtt & .0115 \\
White\_meat\_qtt & .0067 & B2 & .0077 & Egg\_qtt & .0111 \\
Pufa\_omg3 & .0066 & B1 & .0065 & Suppl\_amd & .0108 \\
Bcar & .0063 & COL8A1\_rs140647181 & .0064 & F\_glucose & .0107 \\
COL8A1\_rs140647181 & .0062 & Nf\_glucose & .0058 & CTRB2\_rs72802342 & .0105 \\
Lycop & .006 & CFH\_rs191281603 & .0055 & H\_circ & .01 \\
White\_meat\_freq & .006 & Lycop & .0053 & Legume\_qtt & .0095 \\
Pipe & .0059 & Pufa\_omg3 & .0053 & Cortico & .0092 \\
Nf\_glucose & .0058 & Bcar & .0051 & C2\_rs429608 & .0083 \\
CFH\_rs191281603 & .0053 & C2\_rs144629244 & .0045 & C9\_rs62358361 & .0079 \\
Epa\_dha & .0046 & White\_meat\_freq & .0045 & RDH5\_rs3138141 & .0079 \\
C2\_rs144629244 & .0045 & Cereal\_qtt & .0043 & CFH\_rs187328863 & .0076 \\
CFH\_rs35292876 & .0041 & ACAD10\_rs61941274 & .004 & C3\_rs147859257 & .0075 \\
ACAD10\_rs61941274 & .0038 & C2\_rs181705462 & .004 & Pressure\_dias & .007 \\
Cereal\_qtt & .0036 & White\_qtt & .0037 & COL8A1\_rs140647181 & .005 \\
C2\_rs181705462 & .0035 & CFH\_rs35292876 & .0036 & CFH\_rs121913059 & .0041 \\
White\_qtt & .0035 & Epa\_dha & .003 & CFH\_rs148553336 & .0039 \\
B\_carotene & .0034 & CFI\_rs141853578 & .0028 & CFH\_rs191281603 & .0037 \\
Crypto & .0032 & Crypto & .0027 & C2\_rs181705462 & .0033 \\
Flavoproteins & .0032 & Flavoproteins & .0027 & ACAD10\_rs61941274 & .0032 \\
CFI\_rs141853578 & .0029 & B\_carotene & .0023 & PRLR\_SPEF2\_rs114092250 & .0031 \\
B9 & .0026 & B9 & .0022 & C2\_rs144629244 & .003 \\
Lut\_zea & .0025 & CFH\_rs148553336 & .0021 & Dairy\_qtt & .0029 \\
B3 & .0017 & Lut\_zea & .0015 & Pipe & .0029 \\
CFH\_rs148553336 & .0016 & PRLR\_SPEF2\_rs114092250 & .0014 & CFH\_rs35292876 & .002 \\
PRLR\_SPEF2\_rs114092250 & .0012 & B3 & .0013 & CFI\_rs141853578 & .0015 \\
CFH\_rs121913059 & .0008 & CFH\_rs121913059 & .0006 & Oily\_qtt & .0011 \\
\caption{Ranking of each feature in the EYE-RISK dataset performed by the three CACTUS classification modalities: Degree (CDG), PageRank (CPR), and Probabilistic (CPB).}
\label{tab:ranks}
\end{longtable}

\end{document}